\documentclass[10pt,letterpaper,compsoc,conference]{iiswc25}

\makeatletter
\providecommand{\ps@firstpage}{\ps@IEEEtitlepagestyle}
\makeatother

\usepackage{cite}
\usepackage{amsmath,amssymb,amsfonts}

\usepackage{graphicx}
\usepackage[dvipsnames]{xcolor}
\usepackage[final]{microtype}
\usepackage[italic]{mathastext}
\usepackage{libertine}
\usepackage[T1]{fontenc}
\usepackage{textcomp}
\usepackage[varqu,varl]{zi4}
\usepackage[all]{nowidow}
\usepackage[keeplastbox]{flushend}
\usepackage{fancyhdr}
\usepackage{listings}
\usepackage{xcolor}
\usepackage[ampersand]{easylist}

\lstdefinestyle{bashstyle}{
  language=bash,
  basicstyle=\ttfamily\footnotesize,
  backgroundcolor=\color{gray!10},
  frame=single,
  columns=fullflexible,
  breaklines=true,
  showstringspaces=false
}

\usepackage{float}
\usepackage[hyphens]{url}

\usepackage[ruled,vlined]{algorithm2e}
\DontPrintSemicolon              
\SetKwProg{Function}{Function}{:}{}  

\usepackage{algpseudocode}
\usepackage{makecell}
\usepackage{multirow}
\usepackage{booktabs}
\usepackage{subfiles}
\newcommand{\linebreakand}{
  \end{@IEEEauthorhalign}
  \hfill\mbox{}\par

  \begin{@IEEEauthorhalign}
}

\makeatletter
\NewDocumentCommand{\LeftComment}{s m}{
  \Statex \IfBooleanF{#1}{\hspace*{\ALG@thistlm}}\color{black} \it \(\triangleright\) #2 \color{black} \rm}
\makeatother

\makeatletter
\newcommand\blfootnote[1]{
  \begingroup
  \renewcommand\@makefntext[1]{\parindent 1em\indent ##1}
  \footnotetext{#1}
  \endgroup
}
\makeatother

\begin{document}

\title{Characterizing and Optimizing Real-Time Optimal Control for Embedded SoCs}

\author{
\IEEEauthorblockN{Shengjun Kris Dong\textsuperscript{*}}
\IEEEauthorblockA{University of California, Berkeley\\
krisdong@berkeley.edu}\\[0.3cm]
\IEEEauthorblockN{Minh Nguyen}
\IEEEauthorblockA{University of California, Berkeley\\
minh02@berkeley.edu}
\and

\IEEEauthorblockN{Dima Nikiforov\textsuperscript{*}}
\IEEEauthorblockA{University of California, Berkeley\\
vnikiforov@berkeley.edu}\\[0.3cm]
\IEEEauthorblockN{Vikram Jain}
\IEEEauthorblockA{University of California, Berkeley\\
vikramj@berkeley.edu}\\[0.3cm]
\IEEEauthorblockN{Yakun Sophia Shao}  
\IEEEauthorblockA{University of California, Berkeley\\
ysshao@berkeley.edu}       
\and

\IEEEauthorblockN{Widyadewi Soedarmadji}
\IEEEauthorblockA{Tsinghua University\\
wid4soe@berkeley.edu}\\[0.3cm]
\IEEEauthorblockN{Christopher W. Fletcher}
\IEEEauthorblockA{University of California, Berkeley\\
cwfletcher@berkeley.edu}\\
}

\maketitle
\blfootnote{\textit{*Both authors contributed equally.}}

\thispagestyle{firstpage}
\pagestyle{plain}

\begin{abstract}
Resource-limited robots face significant challenges in executing computationally intensive tasks, such as locomotion and manipulation, particularly for real-time optimal control algorithms like Model Predictive Control (MPC). This paper provides a comprehensive design space exploration to identify optimal hardware computation architectures for these demanding model-based control algorithms. We profile and optimize representative architectural designs, including general-purpose scalar CPUs, vector processors, and specialized accelerators. By characterizing kernel-level benchmarks and end-to-end robotic scenarios, including a hardware-in-the-loop evaluation on a fabricated RISC-V multi-core vector SoC, we present a quantitative comparison of performance, area, and utilization across distinct architectural design points. Our findings demonstrate that targeted architectural modifications, coupled with deep software and system optimizations, enable up to 3.71x speedups for MPC, resulting in up to 27\% system-level power reductions while completing robotic tasks. Finally, we propose a code generation flow designed to simplify the complex engineering effort required for mapping robotic workloads onto specialized architectures. 

\end{abstract}


\section{Introduction}
Robots are increasingly integrated into various sectors, from package delivery and high-precision surgical assistance to autonomous industrial operations. This widespread deployment is empowered by advancements in complex robotic algorithms, encompassing reasoning, sensing, perception, localization, motion planning, and control. These advancements are accompanied by a rapid increase in the complexity of robotic algorithms, which escalates the computational demands of real-time robotics. This poses challenges for execution within stringent latency and power constraints on embedded System-on-Chip (SoC) platforms.

Model predictive control (MPC) is a widely used approach for controlling highly dynamic robotic systems subject to complex constraints. However, its computation demand grows cubically with the robot’s state space and linearly with the prediction horizon. This challenge becomes insurmountable on embedded SoCs as robots evolve into highly dynamic multi-joint rigid body systems. Additionally, control algorithms must run at a frequency of 100-400Hz to meet real-time deadlines in safety-critical applications. Ensuring low-latency computation is crucial in safety-critical applications such as obstacle avoidance and navigation.

Traditional microcontrollers and embedded SoCs either lack the computational capacity to meet such latency requirements or consume excessive energy, thus becoming impractical for battery-powered edge devices.  As robots become more complex and miniaturized, onboard resources, such as compute power, energy, memory storage, and bandwidth, become increasingly constrained. Furthermore, the combination of computation-demanding algorithms, real-time requirements, and resource constraints makes it increasingly impractical to rely on a single hardware architecture to meet all demands on an embedded SoC. To address this issue, modern embedded robotic SoCs, such as the NVIDIA Xavier and Qualcomm Snapdragon processors, increasingly feature diverse, heterogeneous functional units, including CPUs, GPUs, and specialized accelerators. 

Recognizing the inherent complexity of heterogeneous computing, this paper presents a thorough evaluation of hardware architectures suitable for modern robotic systems. Our central thesis is that a systematic, quantitative comparison of foundational architectural paradigms among scalar, vector, and matrix cores is necessary. This exploration extends beyond surface-level performance metrics to analyze the hardware-software interactions that ultimately govern efficiency. The specific contributions of this work are as follows:
\vspace{-3pt}
\begin{enumerate}
  \item A comprehensive design space exploration across scalar, vector, and matrix architectures for real-time robotic optimal control.
  \item Efficient software mapping across architectures, demonstrating that static software scheduling can achieve substantial performance gains of up to a 3.71$\times$ speedup for robotic control algorithms.
    \item Experimental validation of optimizations, with hardware-in-the-loop evaluation on a fabricated RISC-V multi-core vector SoC \cite{jain2025cygnus}, resulting in system-level power reductions of up to 27\% in robotic tasks.

  \item Development of a code generation flow to streamline the mapping of robotic workloads onto the profiled architectures, significantly reducing engineering overhead.

\end{enumerate}

\section{Background and Related Work}
\label{arch}

\subsection{Characterization and Benchmarking}
To bridge the gap between algorithmic demands and hardware capabilities in robotics, prior work has characterized robot workloads on diverse platforms. RoWild \cite{10.1145/3626774,9804645} profiles six end-to-end robotic applications across perception, planning, and control, evaluating systems from embedded Jetson Nano to server-class Xeon CPUs and Titan X GPUs. It finds that general-purpose architectures poorly match robotics workloads, as current vector mappings suffer from inefficient data movement and memory access, motivating the research into improved scheduling and data layouts that we pursue here. RobotPerf \cite{mayoral2024robotperf} provides a benchmarking framework for robotic workloads; this work primarily focuses on methodologies for standardizing robotic benchmarks, rather than characterizing and optimizing the workloads themselves. MavBench \cite{boroujerdian2018mavbench} instead studies system-level effects of robot applications, focusing on CPU workloads. Other work has targeted aerial vehicles, using roofline models to analyze UAV bottlenecks \cite{Krishnan_2022_Roofline} and studying the impact of compute on MAV mission time and energy consumption.

We observe that this prior characterization work targets existing, off-the-shelf hardware (e.g., Jetson Nano, Intel Xeon) and summarizes general bottlenecks without conducting detailed characterization across various architectural designs. Furthermore, while prior work such as RoS\'E \cite{rose} characterizes the performance of robot workloads across various RTL designs, it has limited exploration of software mapping techniques. This reveals a research gap for a focused, quantitative design space exploration that directly compares the foundational architectural paradigms of vector processing and systolic arrays for non-DNN robotic workloads on customized SoC platforms. This paper bridges that gap by conducting fine-grained design space exploration and hardware-software co-optimization for real-time control workloads on customized SoCs. 

\subsection{Architectural Design Points in Robotics}

Robotics algorithms have been mapped to a variety of hardware backends. CPUs are the most commonly used general-purpose computing platforms. Since CPUs are designed to handle a broad range of kernels, cheap CPUs, especially microcontrollers, were traditionally used for simple rule-based robotic kernels such as proportional-integral-derivative (PID) control solvers. However, microcontrollers fail to meet the computational demand for real-time robotic tasks with scaled-up models and DNN-based algorithms. This is mainly because microcontrollers are slow, with a typical compute frequency of 200 MHz. One solution to maximize a CPU's available compute is to exploit the out-of-order and superscalar architecture to drive instruction-level parallelism. However, the performance margin comes at the cost of area overhead and power consumption. An out-of-order and superscalar CPU can achieve up to 100 GFLOP/Sec \cite{chen2024design}. However, its power efficiency is less than 1 GOP/J, which is usually not compatible with power-constrained embedded devices \cite{wang2018survey}.

Vector extensions enable CPUs with lightweight front-ends to surpass the performance of out-of-order and superscalar CPUs, while dramatically simplifying implementation by eliminating excess control logic and exploiting data-level parallelism. Various vector architectures efficiently support scalar, general matrix-matrix multiplication (GEMM), and general matrix-vector multiplication (GEMV) operations and can fall back to the scalar core to support non-standard robotic kernels. These range from traditional long-vector machines \cite{russell1978cray} to packed SIMD (P-SIMD) machines that utilize VLIW instruction encodings, \cite{kozyrakis2002vector}, to contemporary short-vector machines \cite{saturn}.

Another architecture that utilizes data-level parallelism, commonly used for DNN workloads, is a systolic array. These are highly efficient for large, batched GEMM operations. However, since most systolic arrays are tailored explicitly for DNN workloads, their direct application to non-DNN robotics tasks can be challenging. Domain-specific robotic accelerators can significantly boost efficiency for specific robotics workloads; however, their development incurs high engineering costs\cite{8416849, li2019fpga, chretien2016gpu, liang2018gpu, murray2016microarchitecture, lian2018dadu, suleiman2019navion, plancher2021accelerating, Neuman_2021, Neuman_2023, Krishnan_2022}. Both types of architectures require a non-standard programmable interface and programming flow, which makes adoption challenging. Furthermore, their fixed functionality leads to a relatively short lifespan as they can quickly become obsolete when unable to keep pace with the evolving robotic algorithms.

Graphics Processing Units (GPUs) exploit massive data- and thread-level parallelism, making them well-suited for matrix-heavy workloads, graphics, and non-real-time robotics tasks like motion planning or simulation. In contrast, our control workloads involve small tensors (4–150 elements) with sequential dependencies, which cannot leverage GPGPU-scale parallelism. GPUs also face size, weight, and power (SWaP) limitations: their large die area and power draw (often 10-100W+) are impractical for embedded robotic platforms. Thus, we do not target GPUs in our analysis, though they remain attractive for nonlinear MPC (e.g., MPCGPU \cite{plancher2024mpcgpu}) or highly parallel planning workloads (e.g., CuRobo \cite{sundaralingam2023curobo}).

From a high-level perspective, we observe that each architecture offers distinct advantages in terms of computational efficiency by leveraging various forms of parallelism: data-level, instruction-level, and thread-level parallelism to achieve the computational efficiency required for specific types of robotic tasks with different area and power costs. Moving forward, our focus will be on three principal architectural categories: general-purpose CPUs, vector machines, and domain-specific accelerators. These architectures are strategically selected to meet the diverse computational needs of robotic systems within stringent area and power constraints.

\section{Workload Overview}
\label{workloads}

Traditional MPC algorithms are computationally intensive, incorporating dense linear algebra kernels such as GEMV and GEMM, as well as domain-specific algorithms such as Cholesky decomposition and Riccati recursion, which further increase the computational load. Specifically, we profile TinyMPC \cite{tinympc}, a state-of-the-art embedded MPC algorithm as an example, and measure the impact of hardware architectures on kernel performance and end-to-end workloads.

\subsection{Workload Structure}

\begin{figure}
\centering
\includegraphics[width=\linewidth]{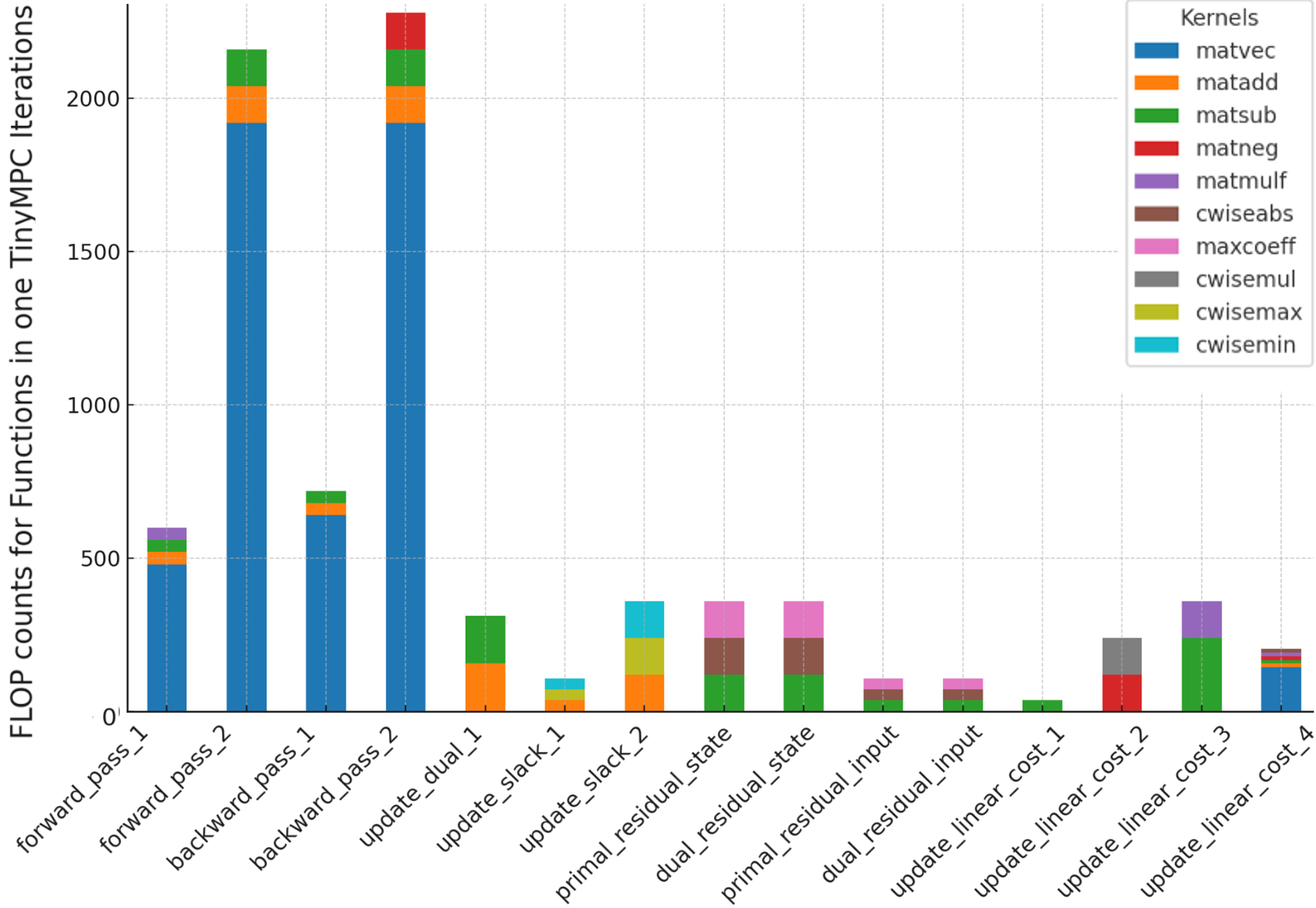}
\caption{FLOP Breakdown of TinyMPC Kernels}
\label{fig:kernel-breakdown}
\end{figure}

The functions of TinyMPC can be broadly classified into three categories: iterative operations with data dependencies, element-wise operations on vectors, and global reductions. These are shown in Algorithms 1, 2, and 3, respectively, with a more detailed kernel breakdown in Figure \ref{fig:kernel-breakdown}.

\begin{algorithm}
\small
\label{alg:iter}
\caption{\small Iterative Operations in TinyMPC}
\LinesNumbered

  \Function{forward\_pass\_1($i$)}{
    u[i] $\gets$ $-Kinf \cdot x[i] - d[i]$\;
  }

  \Function{forward\_pass\_2($i$)}{
    x[i+1] $\gets$ $Adyn \cdot x[i] + Bdyn \cdot u[i]$\;
  }

  \Function{backward\_pass\_1($i$)}{
    d[i] $\gets$ $Quu\_inv \cdot \bigl(Bdyn^\top \cdot p[i+1] + r[i]\bigr)$\;
  }

  \Function{backward\_pass\_2($i$)}{
    p[i] $\gets$ $q[i] + AmBKt \cdot p[i+1] - Kinf^\top \cdot r[i]$\;
  }

  \Function{update\_linear\_cost\_4()}{
    p[N-1] $\gets -Xref[N-1]^\top Pinf\,(vnew[N-1] - g[N-1])$\;
  }

\end{algorithm}

\begin{algorithm}
\caption{\small Elementwise Operations in TinyMPC}
\small
\label{alg:stripmine}

  \Function{update\_slack\_1()}{
    $z_{\text{new}} \gets u + y$\;
    $z_{\text{new}} \gets \min(u_{\max},\;\max(u_{\min},\;z_{\text{new}}))$\;
  }

  \Function{update\_slack\_2()}{
    $v_{\text{new}} \gets x + g$\;
    $v_{\text{new}} \gets \min(x_{\max},\;\max(x_{\min},\;v_{\text{new}}))$\;
  }

  \Function{update\_dual\_1()}{
    $y \gets y + u - z_{\text{new}}$\;
    $g \gets g + x - v_{\text{new}}$\;
  }

  \Function{update\_linear\_cost\_1()}{
    $r \gets -\rho\bigl(z_{\text{new}} - y\bigr)$\;
  }

  \Function{update\_linear\_cost\_2()}{
    $q \gets -\bigl(X_{\text{ref}}\cdot Q\bigr)$\;
  }

  \Function{update\_linear\_cost\_3()}{
    $q \gets q - \rho\bigl(v_{\text{new}} - g\bigr)$\;
  }

\end{algorithm}

\begin{algorithm}[t]
\caption{\small Global Reductions in TinyMPC}
\small
\label{alg:reduce}

  \Function{primal\_residual\_state()}{
    primal\_residual\_state $\gets \max\bigl(\lvert x - vnew\rvert\bigr)$\;
  }

  \Function{dual\_residual\_state()}{
    dual\_residual\_state $\gets \rho\,\max\bigl(\lvert v - vnew\rvert\bigr)$\;
  }

  \Function{primal\_residual\_input()}{
    primal\_residual\_input $\gets \max\bigl(\lvert u - znew\rvert\bigr)$\;
  }

  \Function{dual\_residual\_input()}{
    dual\_residual\_input $\gets \rho\,\max\bigl(\lvert z - znew\rvert\bigr)$\;
  }

\end{algorithm}

\subsection{Programming Interface and Libraries}
To create a unified platform for comparing the performance of differing architectures for robotic and DSP applications, we create a C library of commonly used operators, \texttt{matlib}\footnote{\url{ https://github.com/wid4soe/matlib}}.

\texttt{matlib} provides a lightweight interface to various linear algebra operations similar to Eigen \cite{eigenweb}, a C++ header-only library. \texttt{matlib} was used to write reference implementations for each backend; however, to achieve full performance, hand-tuned implementations required optimizations across the abstractions provided by these function definitions.

\section{Characterization and Optimization}
To identify the optimal design choice for real-time robotic control algorithms, we conducted a multifaceted design space exploration and evaluation, encompassing the selection of target architectures, hardware-software co-optimization, the development of automated code generation flows, and an evaluation framework that included hardware-in-the-loop experiments. 

\begin{figure}[t]
\centering
\includegraphics[width=\linewidth]{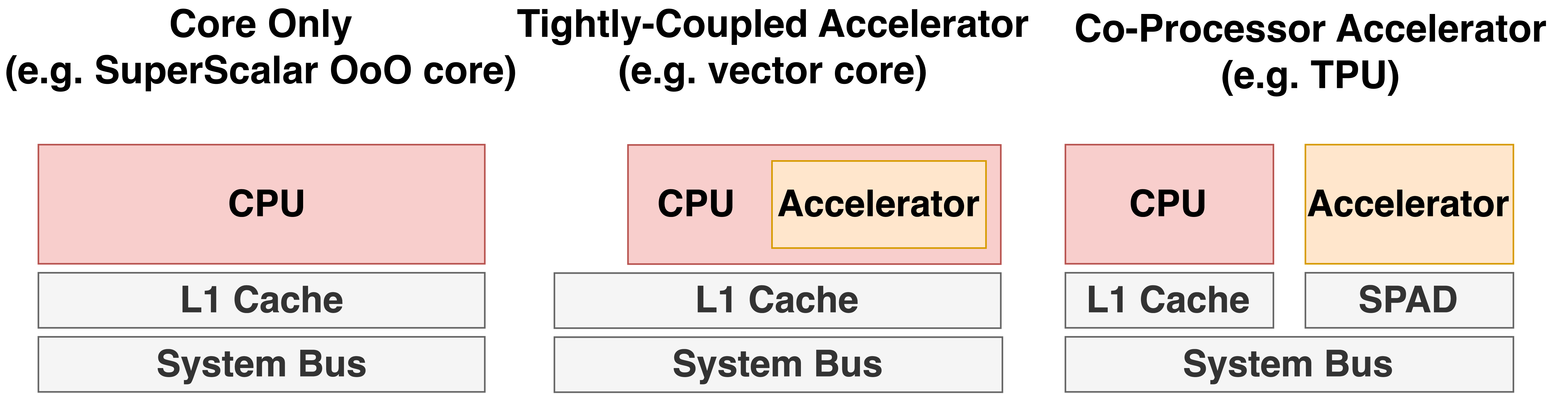}
\caption{Categories of Target Hardware Architectures to be Profiled and Evaluated: CPU, tightly integrated accelerators, and decoupled co-processors.}
\label{fig:arch}

\centering
\end{figure}

We target three primary categories: general-purpose CPUs, vector machines, and domain-specific accelerators as depicted in Figure~\ref{fig:arch}: For general-purpose CPUs, we evaluate a variety of different configurations of RISC-V CPU cores, including simple in-order core Rocket \cite{Rocket-RISCV-2016}, superscalar in-order core Shuttle \cite{ucb_bar_shuttle}, and superscalar out-of-order BOOM core \cite{zhaosonicboom}. For vector machines, we evaluate Saturn-V\cite{Zhao:EECS-2024-215}, an implementation of the RISC-V vector extension (RVV 1.0) with a short-vector microarchitecture.  For domain-specific accelerators, we profile a systolic array generated using Gemmini\cite{genc2019gemmini}.  

Through careful characterization from the prior section, we identify a critical performance barrier: the inherent mismatch between the design of the target hardware schedulers (e.g., sequencers in vector cores, reorder buffers in systolic arrays) and the specific computational patterns of robotics control workloads. In this section, we investigate the friction between hardware scheduling mechanisms and the control flow of robotic algorithms, and develop highly optimized software mapping accordingly.

\subsection{Optimizations for Vector Cores}
\begin{figure}
    \centering
    \includegraphics[width=\linewidth, trim=0cm 0cm 0cm 0cm, clip]{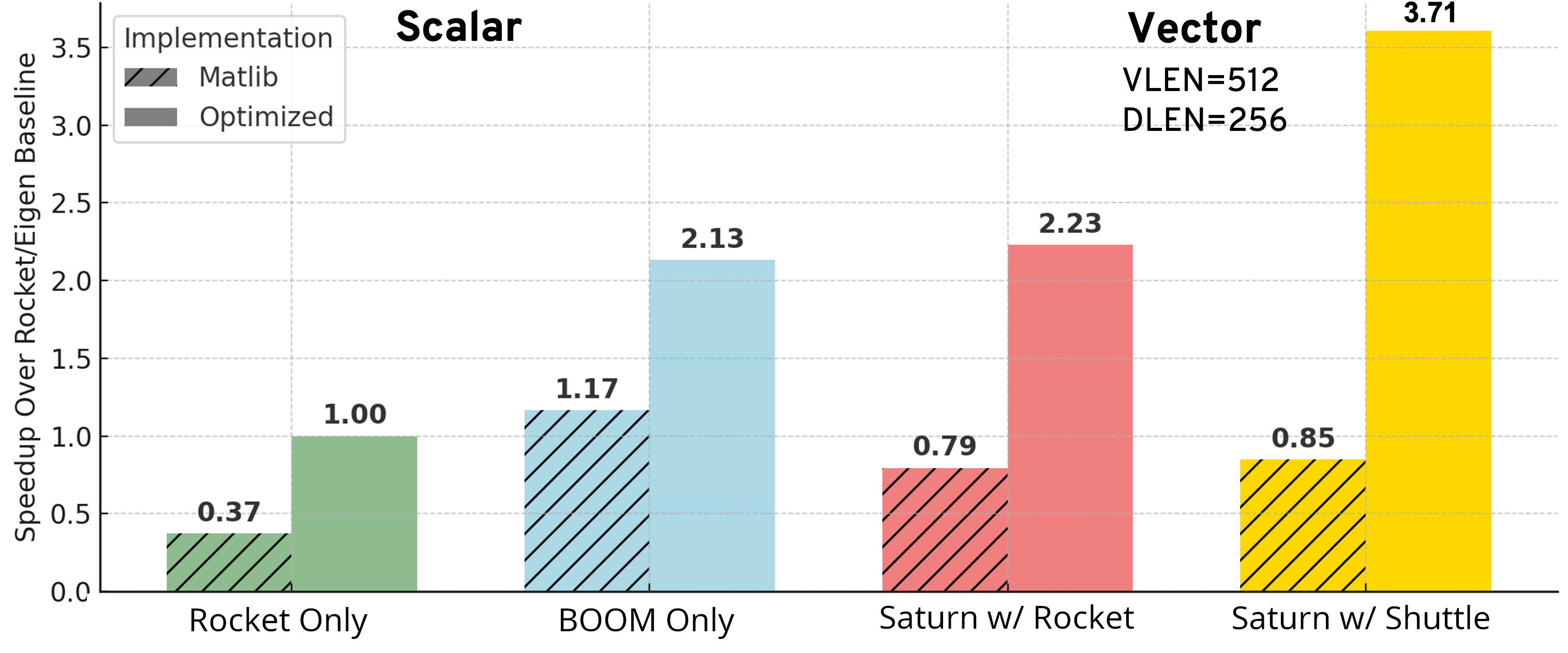}
    \caption{Out-of-box application of \texttt{matlib} vs hand-optimized vectorized TinyMPC}
    \label{fig:matlib-perf}
\end{figure}

To initially accelerate TinyMPC, we utilize a library-based approach. For every \texttt{matlib} function that is used within the solver, we write a vectorized implementation using RISC-V vector extension (RVV) intrinsic support from \texttt{gcc}. However, as shown in Figure~\ref{fig:matlib-perf}, although the vectorized \texttt{matlib} code has a meaningful speedup over scalar \texttt{matlib} code running on Rocket, highly optimized scalar code using Eigen still outperforms the Saturn implementation, necessitating further optimizations.

\subsubsection{Hardware and Software Loop Unrolling}
\begin{figure}
    \centering

    \includegraphics[width=1.0\linewidth, trim=0cm 0cm 0cm 0cm, clip]{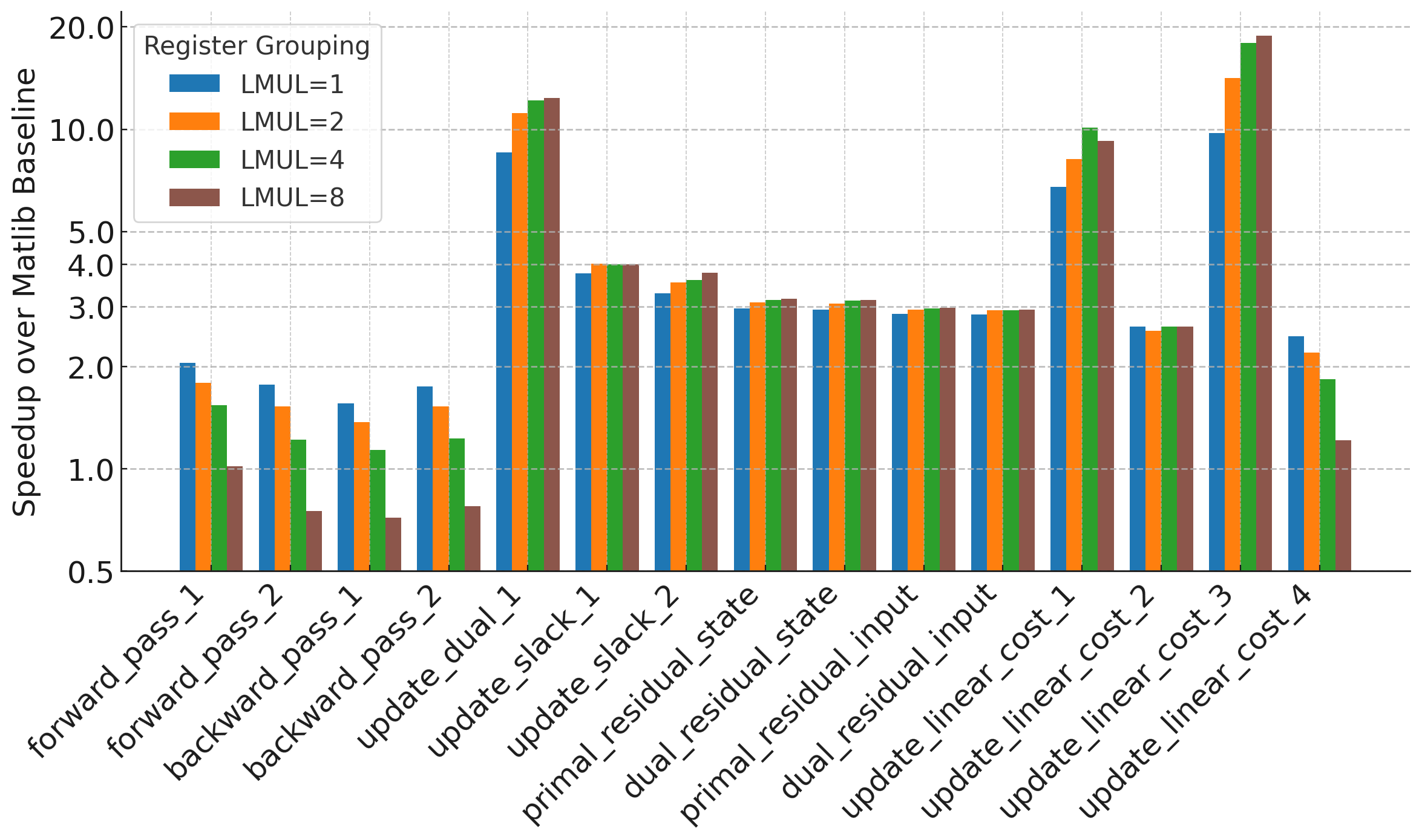}
    \caption{Performance of TinyMPC using Saturn with varying sizes of LMUL }
    \label{fig:lmul}
\end{figure}

A key feature of the RVV ISA is register grouping via the \texttt{LMUL} field, which unrolls contiguous vector operations in hardware and reduces the need for software unrolling. As shown in Figure~\ref{fig:lmul}, \texttt{LMUL} improves elementwise performance in \texttt{matlib}, but degrades TinyMPC’s iterative kernels (\texttt{backward\_pass}, \texttt{forward\_pass}). These rely on serial GEMV and vector additions that cannot map to larger registers due to dependencies. Optimized GEMV on Saturn uses \texttt{vfmacc\_vf}, multiplying a scalar against a matrix column, but TinyMPC's small vector sizes (4 and 12) only partially fill 512-bit registers, limiting \texttt{LMUL} benefits. In these cases, aggressive software loop unrolling better exploits scalar variation across instructions.

\begin{figure}
    \centering

    \includegraphics[width=\linewidth, trim=0cm 0cm 0cm 0.0cm, clip]{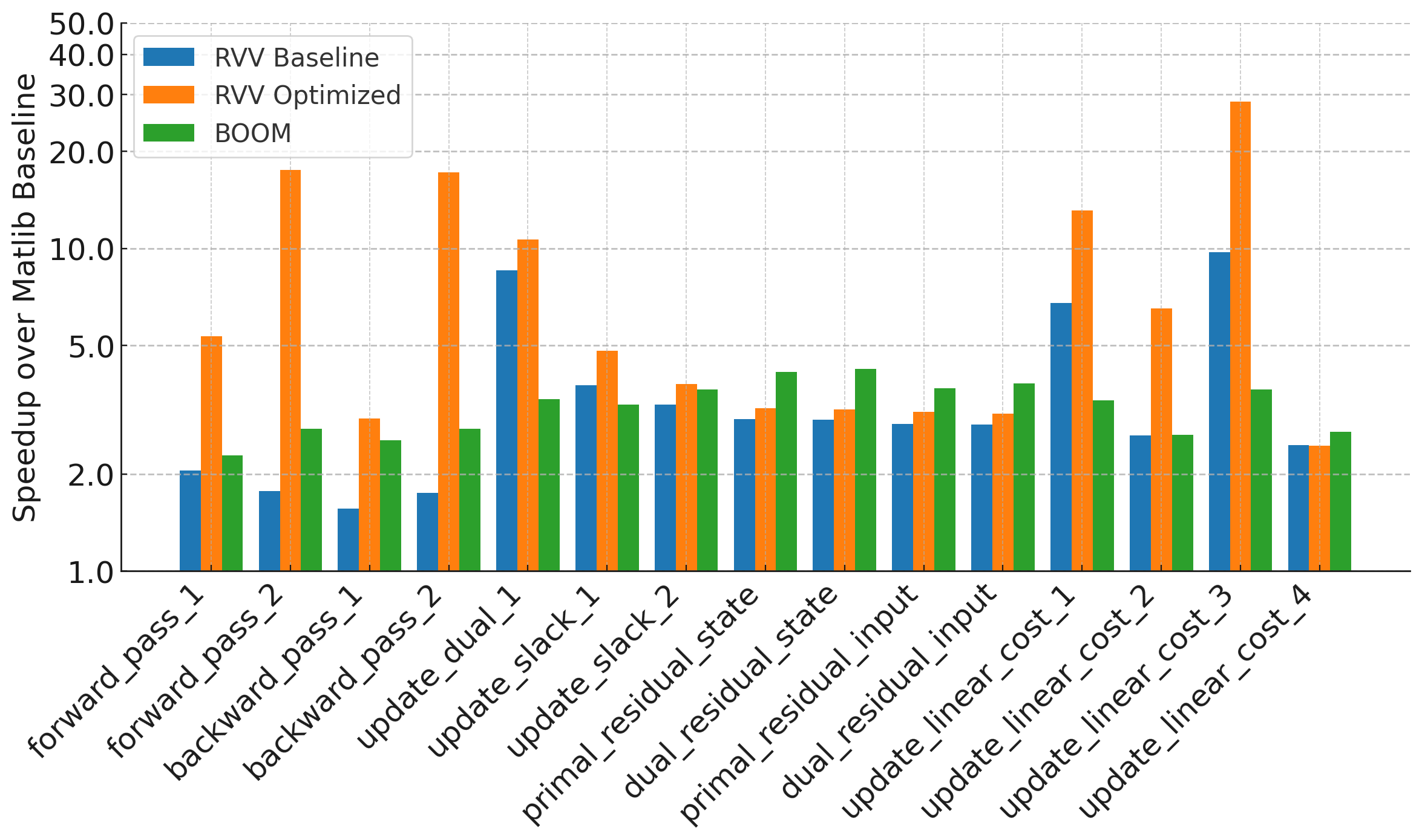}
    \caption{Library vs Fused-Operator Speedup on Rocket-driven 512V256D Saturn
}
    \label{fig:saturn-fusion}
\end{figure}

\subsubsection{Operator Fusion}
Another downside of optimizing library-based code for control applications is that the function boundaries prevent the fusion of operators, thereby limiting register reuse. Every time a \texttt{matlib} function is called, data are explicitly written back to memory using RVV store intrinsics and must be subsequently loaded back to registers in future invocations. This increases the instructions that the frontend must serve to the vector units and adds additional memory latency. Instead, we write optimized implementations of the TinyMPC functions, fusing \texttt{matlib} operators by keeping temporary values within the vector registers until they need to be written back to the TinyMPC workspace. The impact of these optimizations, along with software unrolling, is depicted in Figure~\ref{fig:saturn-fusion}. 

\subsection{Optimizations for Systolic Arrays}

\subsubsection{Static Scheduling of Gemmini Kernels} Static mapping of kernels involves pre-calculating and optimizing the allocation of data and tasks to various processing elements within the hardware. In scenarios involving fixed-size operations commonly found in MPC, dimensions, tiling, and indexing can be determined without the need for dynamic memory allocation. This allows for the entire computational workflow to be streamlined, as arguments for instructions such as moving data in (i.e. \texttt{mvin}) and out (i.e. \texttt{mvout}) of the scratchpad and computation (i.e. \texttt{execute}) can be computed statically. This approach significantly reduces the computational overhead associated with dynamically calculating these parameters during runtime. This reduces the overhead on the CPU driving the accelerator, improving the accelerator's utilization due to the improved throughput of accelerator instructions.

\subsubsection{Reduction of Redundant Configuration Operations} We can additionally optimize for the removal of redundant accelerator operations. This includes unnecessary configuration commands, excessive fencing operations, and redundant data movement already present in the system. For example, we can reuse the accelerator configuration when executing a sequence of operations with the same matrix dimensions. In addition to reducing overall instruction count, this also enables more efficient usage of accelerator compute resources.

\begin{figure}
    \centering
    \includegraphics[width=1.1\linewidth]{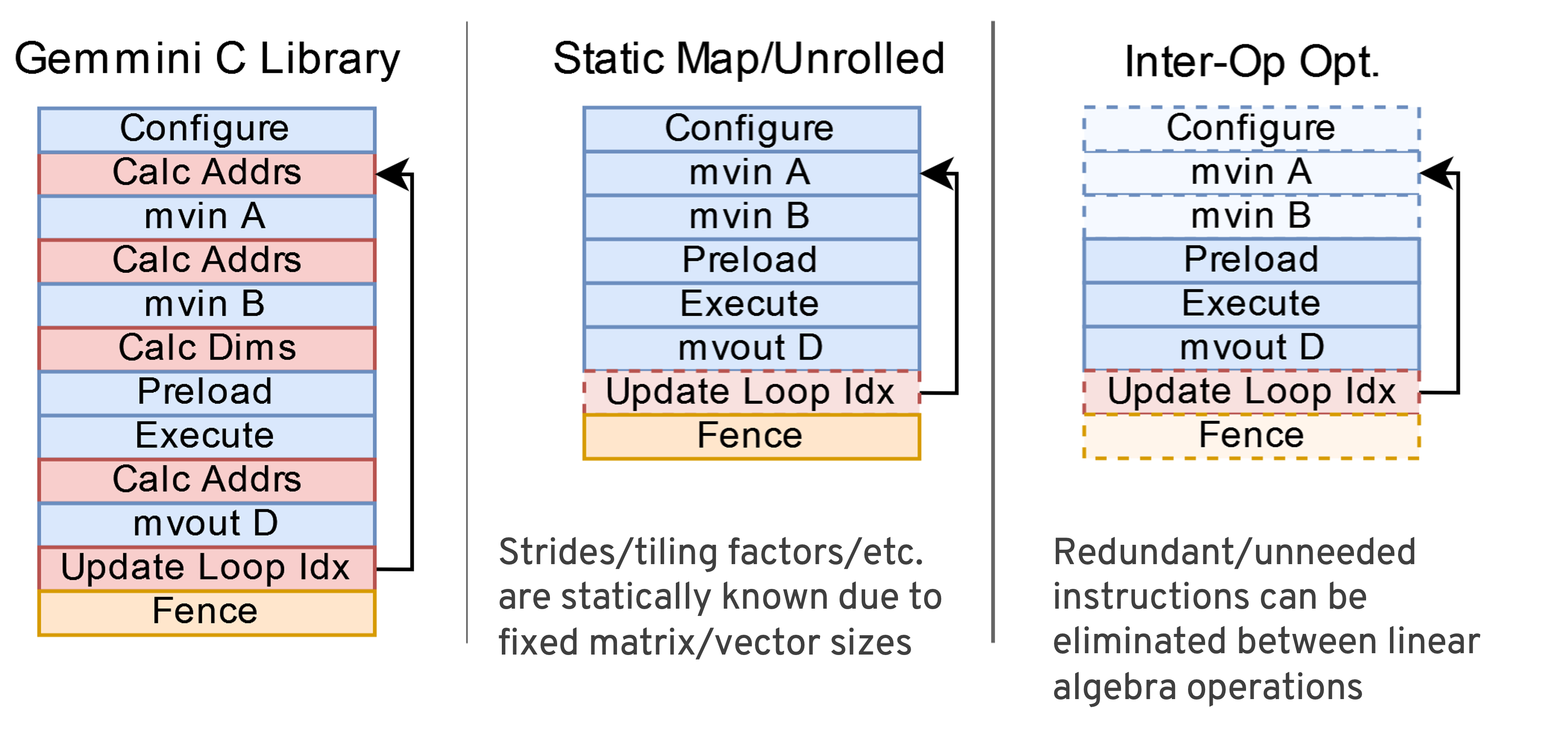}
    \caption{Optimizing Gemmini software mapping with loop unrolling and static mapping}
    \label{fig:enter-label}
\end{figure}

\subsubsection{Use of Gemmini's Fine-Grained ISA}

Gemmini's programming interface uses RoCC, an extension to the RISC-V instruction set \cite{gemmini-dac}, and supports both fine-grained instructions (such as individual loads) and CISC instructions. The latter consists of complex operations such as tiled matrix multiplication, which are sequenced into fine-grained instructions in hardware.

Gemmini's CISC instruction set is typically used for most machine-learning applications to achieve high mesh utilization. However, CISC instructions require multiple RoCC instructions for accelerator configuration before execution can begin. Constructing these RoCC instructions can take multiple cycles due to the bit-shifting that needs to be performed by the scalar CPU to construct the RoCC arguments. Due to the smaller matrix sizes found in TinyMPC, CISC instructions are only mapped to 2-6 fine-grained instructions, resulting in a negligible benefit from hardware instruction sequencing. Furthermore, CISC instructions require the input operands to be stored in memory, preventing additional optimizations from reusing data on Gemmini's scratchpad.

However, using the fine-grained ISA requires improved instruction throughput from the CPU, driving the RoCC interface to achieve high utilization of Gemmini. To address this, we aggressively unroll code and perform indexing and address calculation at compile-time using static mapping to minimize the overhead of constructing RoCC instructions.

\subsubsection{Scratchpad-Resident Linear Algebra Operations}
To support data-dependent operations on Gemmini, modifying the typical DNN kernel mapping for Gemmini is necessary. Typically, inputs to an operation are loaded into Gemmini's scratchpad before they are used as inputs to tiled operations, accumulating in Gemmini's accumulator memory. These results are then stored back in DRAM before being reused in subsequent operations. However, when performing operations that take several cycles on Gemmini, this overhead severely limits performance. In addition to the latency induced by data movement, explicit fence instructions must be inserted between Gemmini store and load instructions, as Gemmini's ROB does not track data RAW hazards across memory operations. In our experimentation, this can introduce up to 600 cycles of stalling upon a fence instruction within the iterative phase of TinyMPC.
\begin{figure}
    \centering
    \includegraphics[width=\linewidth, trim=0cm 0cm 0cm 0cm, clip]{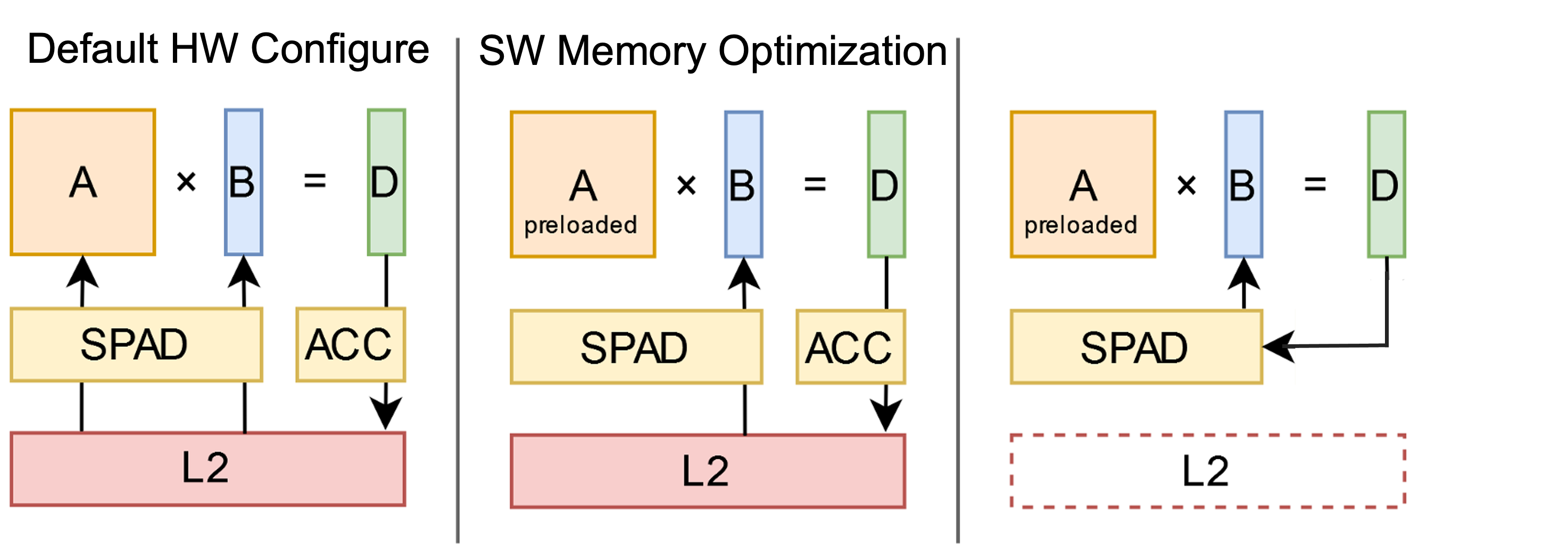}
    \caption{Optimizing Gemmini's Memory for Scratchpad-Resident Workloads}
    \label{fig:enter-label}
\end{figure}

To avoid this issue, we perform several optimizations to our mapping. First, we load all matrices used by TinyMPC onto the first bank of the scratchpad, along with several utility matrices, including the identity and negative identity matrices. Secondly, intermediate results computed by Gemmini are directly written to the scratchpad, allowing them to be immediately reused in subsequent operations. One downside of this approach is that writing to the scratchpad instead of the main memory prevents the use of Gemmini's output scaling pipeline, which can help perform fused scaling/GEMM operations. To compensate for this, we allocate additional utility matrices onto the scratchpad, which are commonly used scalar multiples of the identity matrix.

Although it is theoretically possible to perform most of TinyMPC's kernels fully resident on the scratchpad, we only perform this optimization for the iterative passes. The reason for this is that when performing GEMV operations using Gemmini's original hardware, vectors must be stored in a single column of the destination scratchpad, resulting in inefficient loads when performing element-wide operations, as only one element is loaded per cycle. However, future architectural enhancements to Gemmini, such as hardware GEMV support, would allow vectors to be stored and packed across scratchpad rows, addressing this issue.

\subsubsection{Output Stationary Dataflow and Elimination of Accumulator Memory}
Storing the outputs of Gemmini's compute operations in the scratchpad prevents unnecessary data movement to memory. However, this also prevents the use of hardware accumulators to perform tiled operations on the same partial outputs. To address this, this work uses a Gemmini configuration with an output-stationary dataflow. Unlike machine learning workloads, there is not enough significant reuse of weights to benefit from a weight-stationary dataflow, and accumulating within the mesh's PEs eliminates the need for an explicit accumulator memory. Furthermore, although Gemmini does not support using CISC instructions for the output stationary dataflow, because CISC instructions are not used for TinyMPC, there is no drawback to using the output stationary dataflow from a programming interface perspective.

\subsubsection{Secondary use of DNN Activation Functions and Pooling}
Although linear operations such as vector addition and subtraction, scalar multiplication, and GEMV can be performed solely using Gemmini's mesh, other operations, such as absolute value and min/max, cannot be performed. The absolute value is required to perform the element-wise operations in TinyMPC on Gemmini. However, as depicted in Equation~\ref{eq:abs}, the absolute value can be implemented using ReLU, an activation function natively supported by Gemmini. This enables Gemmini to compute strip-mining operations fully, leveraging Gemmini's capability to fuse activation functions with tiled operations on the mesh.
\begin{equation}\label{eq:abs}
    \mathrm{abs}(x) = \mathrm{ReLU}(x) + \mathrm{ReLU}(-x)
\end{equation}
Additionally, ReLU can also be used to clip a value to an upper or lower bound, which is required for updating slack variables in TinyMPC. However, these operations can also be performed using ReLU using Equations~\ref{eq:clipmin} and~\ref{eq:clipmax}.
\begin{equation}\label{eq:clipmin}
    \mathrm{clip_{low}}(x, \mathrm{min}) = \mathrm{ReLU}(x -\mathrm{min}) +\mathrm{min}
\end{equation}
\vspace{-0.3cm}
\begin{equation}\label{eq:clipmax}
    \mathrm{clip_{high}}(x, \mathrm{max}) = -\mathrm{ReLU}(-x+\mathrm{max}) +\mathrm{max}
\end{equation}
Finally, computing the residuals of TinyMPC requires calculating a global maximum across vectors of approximately 100 elements each. While this could potentially be implemented using sequences of ReLU operations, computing a maximum between two arbitrary vectors would require over five operations, achieving comparable performance to simply using a scalar CPU. However, to avoid computing the entire maximum on the scalar core, we utilize Gemmini's hardware support for max-pooling. By using a pool size of 2 when moving out to shared memory, Gemmini can perform a reduction across four scratchpad rows. This can potentially be increased by increasing the hardware pooling dimensions. Although this does not enable reductions within a scratchpad row, it still reduces the maximum reduction that needs to be performed on the CPU by a factor of 4.

\begin{figure}
    \centering
    \includegraphics[width=0.7\linewidth]{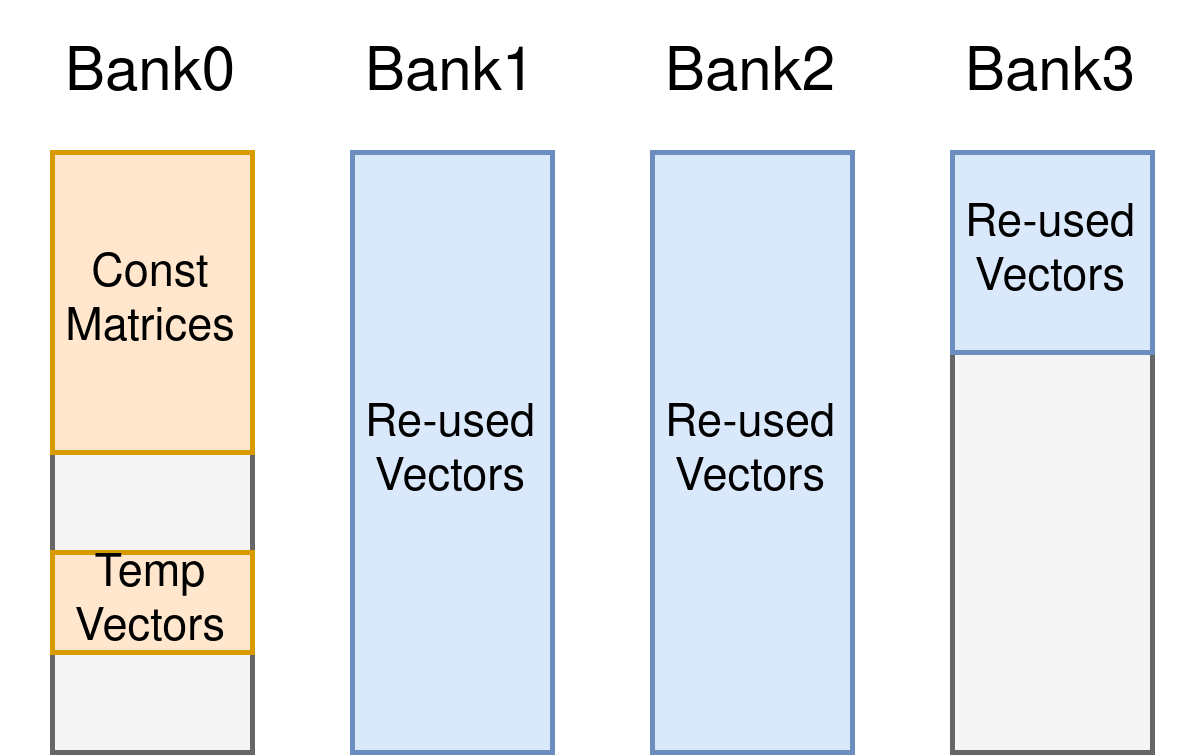}
    \caption{Mapping of the TinyMPC Solver Workspace to Gemmini Scratchpad}
    \label{fig:enter-label}
\end{figure}

\subsubsection{Develop Fine-Grained Synchronization Interface}
This optimization aims to reduce synchronization overheads in memory interfaces, which can significantly hamper computational efficiency in parallel processing environments. By streamlining the synchronization processes and enhancing the communication protocols between the CPU and Gemmini, we reduced the number of fence instructions to avoid stalls on the CPU side. This reduces the time wasted waiting for memory accesses and synchronization, improving the overall system responsiveness and throughput.
\begin{figure}
    \centering
    \includegraphics[width=1\linewidth]{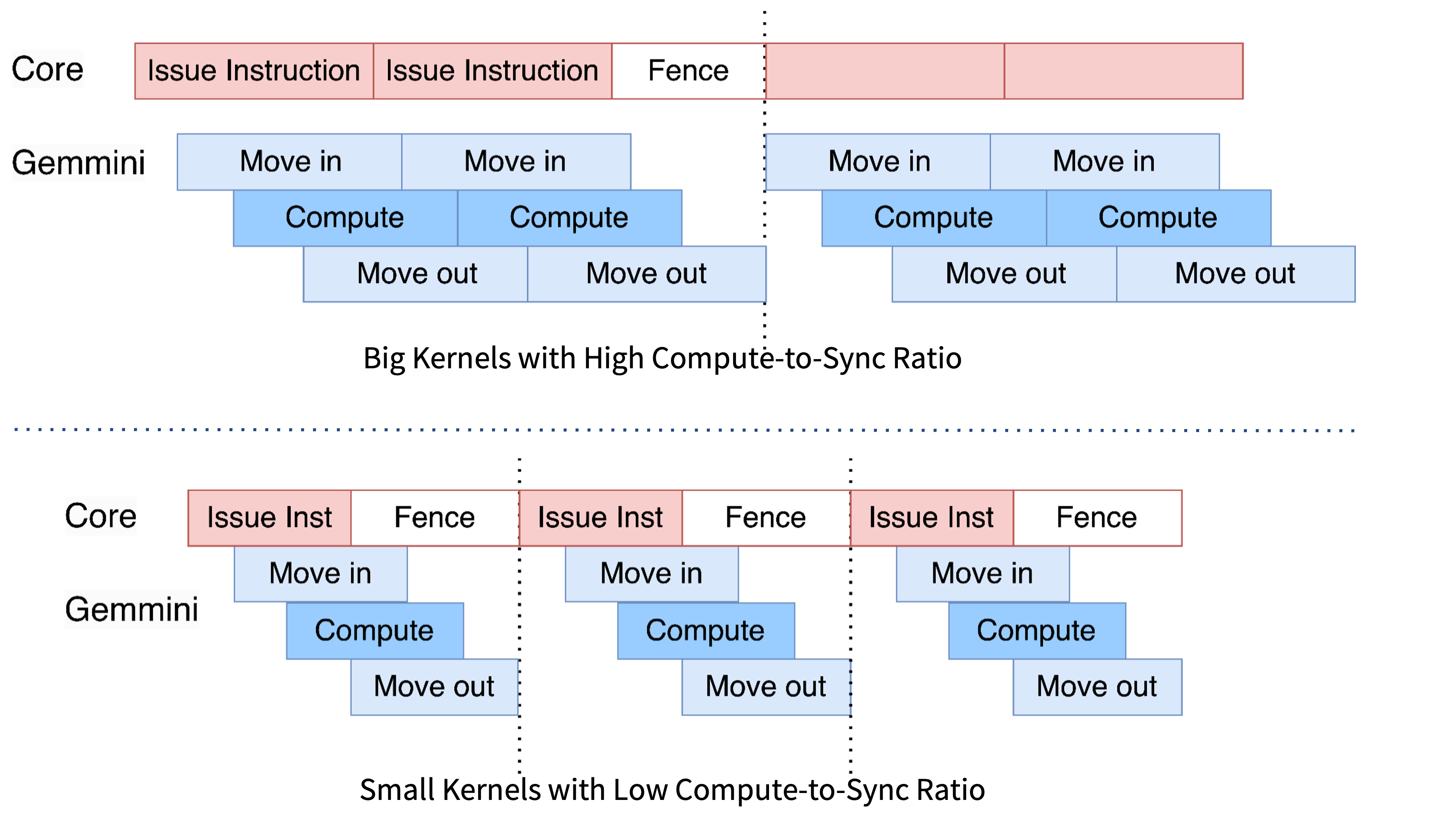}
    \caption{The Impact of Kernel Granularity on CPU-Gemmini Synchronization Overhead}
    \label{fig:enter-label}
\end{figure}

\subsection{Automated Code Generation}
Although we observe significant performance benefits from hand-optimizing our kernels, this requires considerable engineering effort and is not portable to new applications. To address this, we implement a code generation flow in our \texttt{matlib} library.  First, we provide a baseline implementation of an ADMM solver using RVV intrinsics. We then provide an optimization pass that traverses the C abstract syntax tree (AST) to apply customized tiled and batched code unfolding (to reduce overhead from looping, branching, and unnecessary register transfers), as well as automated operator fusion that can minimize register uses for compatible elementwise operations. The library and tools support \texttt{float32} and \texttt{float64} vector and matrix operations, and can use different tile and batch sizes that fit the problem parameters. When solving a quadrotor tracking problem, the baseline CPU version consumes about 11 million cycles, while the baseline vectorized version (without register grouping) consumes 1.35 million cycles, and our automated unrolled and fused version consumes only 0.55 million cycles. This shows that \texttt{matlib} offers a simple-to-program yet performant way to vectorized operations on platforms. Furthermore, compared to traditional libraries, which are optimized for SIMD ISAs, our approach preserves RVV programming practices such as the use of variable-length vectors. Finally, by optimizing sequences of \texttt{matlib} invocations at compile time, we can automate operator fusion, which is not possible with libraries that have individually optimized operators.

\section{Evaluation}
\label{evaluation}

\begin{figure}[t]
\centering
\includegraphics[width=1.0\linewidth]{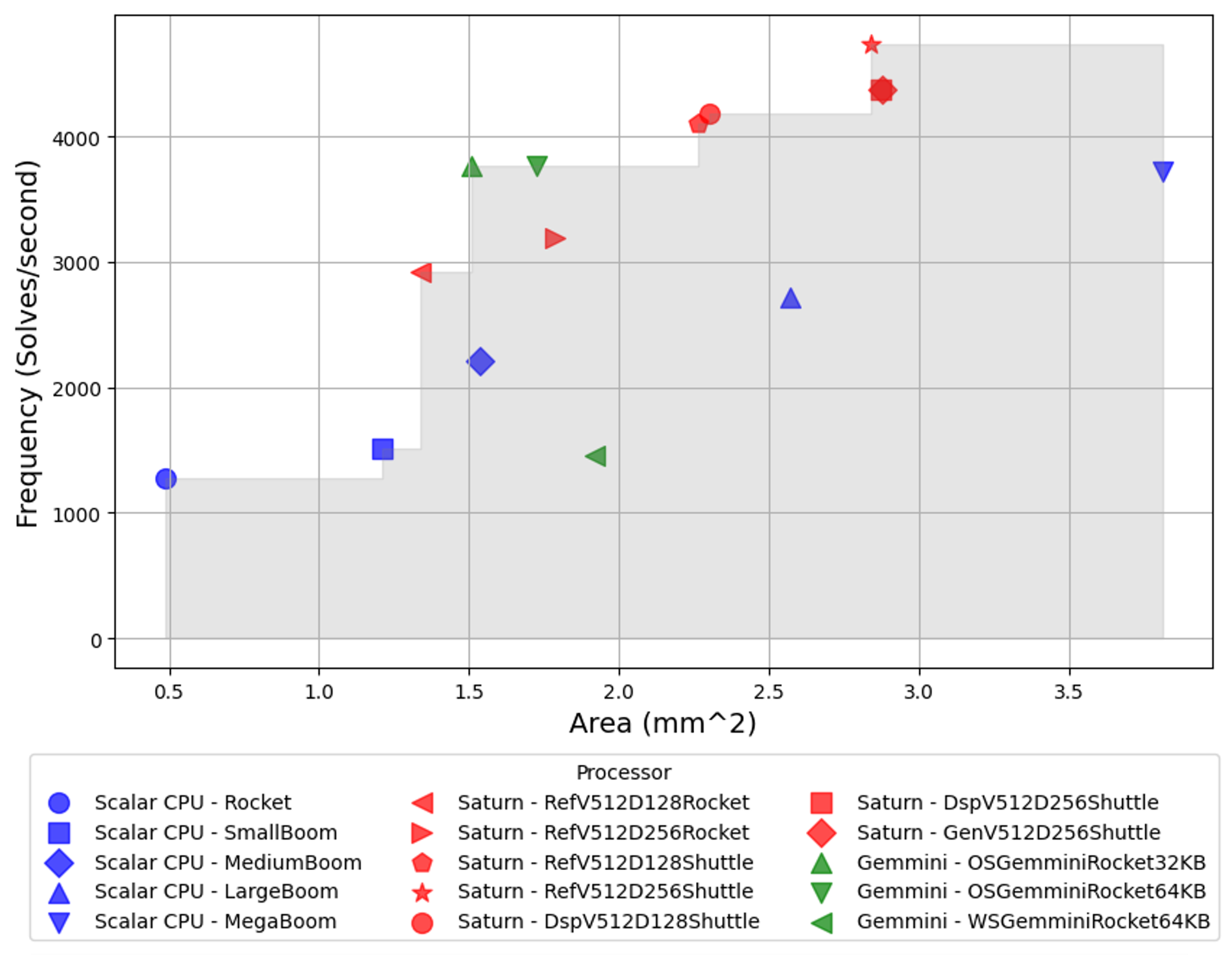}
\caption{Superscalar, Vector, and Systolic Performance vs Area Tradeoffs}
\label{fig:pareto}
\vspace{-0.5cm}
\centering
\end{figure}
\subsection{Algorithm-level Performance Evaluation}

To compare the performance of hardware backends on a real robotic workload, we evaluated the performance of all hardware designs in computing TinyMPC. For this section, we use post-synthesis area results generated by the ASAP7 toolkit \cite{asap7}, while performance numbers are obtained from RTL simulation. Figure~\ref{fig:pareto} plots each hardware configuration using its area, as well as the average frequency at which it can execute the ADMM solver used in TinyMPC. Furthermore, this chart highlights the Pareto-optimal frontier for this application. Reflecting the efficiency gains of switching to specialized architecture compared to general-purpose processors, vector and systolic implementations are optimal outside of highly area-constrained design points.

\subsubsection{Performance Evaluation of Scalar RISC-V CPUs}

Breaking down performance within each category, we profiled representative architecture candidates: RISC-V scalar cores, Saturn RISC-V CPUs with vector extensions, and Gemmini systolic arrays. The RISC-V CPUs included a simple in-order core, a superscalar in-order core, an out-of-order BOOM core, and a high-performance MegaBOOM core with 2 FPUs. For end-to-end control workloads, we use Rocket running optimized scalar Eigen code as a baseline.

The BOOM variants illustrate design trade-offs. Small BOOM has a fetch width of 4, decode width of 1, and three instruction queues (IQs) for MEM, INT, and FP pipelines. Medium BOOM keeps fetch width at 4 but raises decode width to 2, with expanded IQ configurations: MEM (1 issue, 2 dispatch), INT (2 issue, 2 dispatch), and FP (1 issue, 2 dispatch). Large BOOM also fetches 4 and decodes 1, but increases dispatch capacity across MEM, INT, and FP. Mega BOOM scales further to fetch width 8 and decode width 4, with higher issue and dispatch rates across all pipelines.

Scaling up CPU configurations improves end-to-end workload performance, especially with out-of-order and superscalar execution in BOOM. Dedicated IQs for each pipeline enhance instruction-level parallelism, maximizing compute availability. However, these gains incur area and power costs, and CPUs more complex than Small BOOM perform suboptimally, as shown in Figure~\ref{fig:pareto}.

\subsubsection{Performance Evaluation of Saturn}

\begin{figure}[t]
\centering
\includegraphics[width=\linewidth]{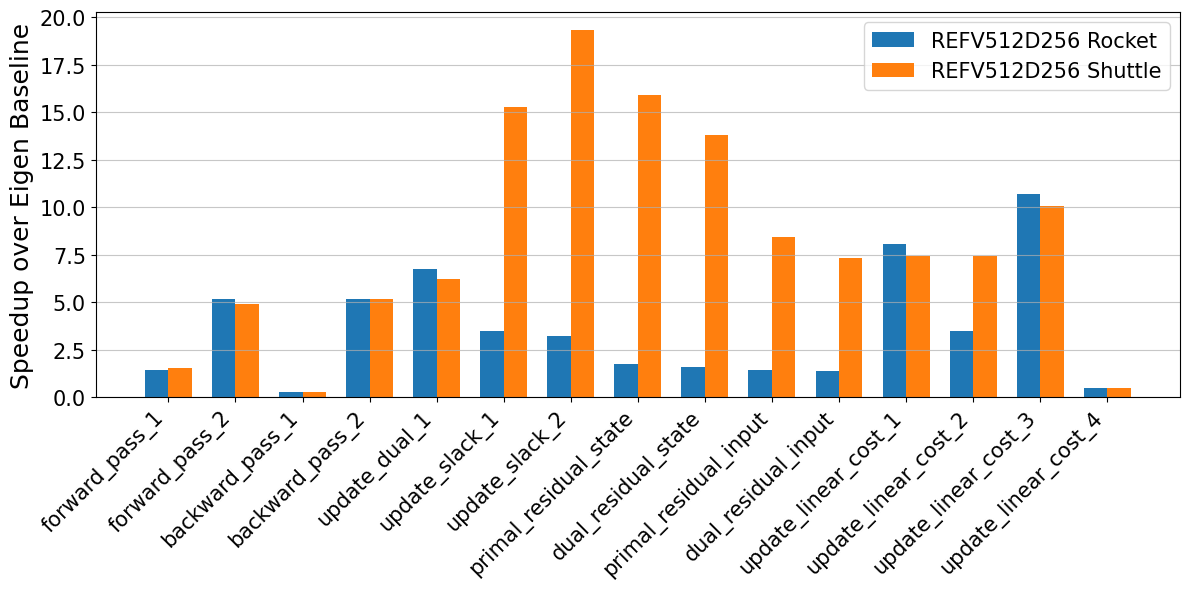}
\caption{Comparison of Kernel Performance on Saturn with a Rocket and Shuttle Frontend}
\label{fig:saturn256_perf}
\vspace{-0.3cm}
\centering
\end{figure}

Saturn shows strong performance across a variety of operations but is particularly effective in tasks like primal and dual state updates, residual calculations, and linear operations based on Figure~\ref{fig:saturn256_perf}, as these kernels have independent elementwise operations on larger tensors.  In these cases, Saturn can fully map operations to vector registers without significant overhead from handling tail cases, and can utilize instruction sequencing and register grouping. However, to achieve high performance, a dual-issue Shuttle frontend is required, as the vector backend underperforms when using a single-issue Rocket core.

\subsubsection{Performance Evaluation of Gemmini}

\begin{figure}[h]
\centering
\includegraphics[width=1\linewidth]{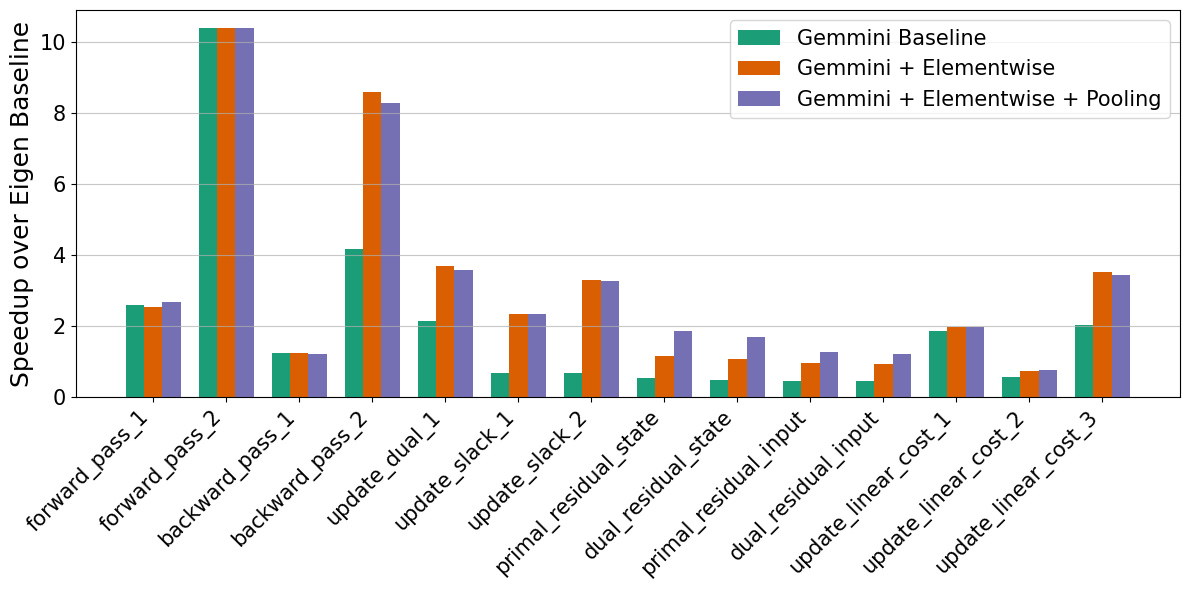}
\caption{Performance of Gemmini (4x4 FP Mesh) on target workload TinyMPC with kernel breakdowns. The Gemmini baseline only uses the mesh for computation, while elementwise and pool signify the use of scaling/activation and pooling engines, respectively.}
\label{fig:Gemmini_perf}
\vspace{-0.1cm}
\centering
\end{figure}

Gemmini's performance peaks in specific operations like forward passes and linear cost updates, as shown in Figure~\ref{fig:Gemmini_perf}, suggesting that its architecture is highly optimized for scenarios where matrix-vector operations are dominant. The significant variability in Gemmini's performance across different operations may indicate its specialized nature, excelling greatly in its niche (matrix operations)  while being less efficient for other types of data processing tasks compared to general-purpose vector processors like Saturn. However, the scaling, activation, and pooling engines intended for AI applications can also be used to accelerate elementwise and reduction operations.

\subsubsection{Comparison Setup}
Furthermore, we conduct an initial comparison between Gemmini and Saturn to better understand each architecture's relative performance and its unique strengths in certain types of operations. The configurations are carefully chosen, such as Gemmini with 4$\times$4 FP Mesh and Saturn V512D512, both driven by Rocket. These configurations ensure that the backends of each architecture have equal numbers of PEs, allowing them to achieve the same ideal FLOPs per cycle. Besides the same computational capacity of the backend of each candidate, choosing the front end to be Rocket for both also allows the instructional throughput to be relatively similar between the two candidates.

At a high level, Saturn shows more uniform and often higher speedup across a broad range of operations, suggesting better general-purpose usability with its vector extensions, especially for varied computational tasks. On the other hand, Gemmini excels in specific tasks tailored to its systolic array architecture, achieving exceptional speedup in matrix-related operations but showing less versatility across a broad spectrum of computational tasks.

\subsubsection{Performance Comparison Across all Architectures}

\begin{figure}[h]
\centering
\includegraphics[width=\linewidth, trim=0.1cm 0.1cm 0.1cm 1.8cm, clip]{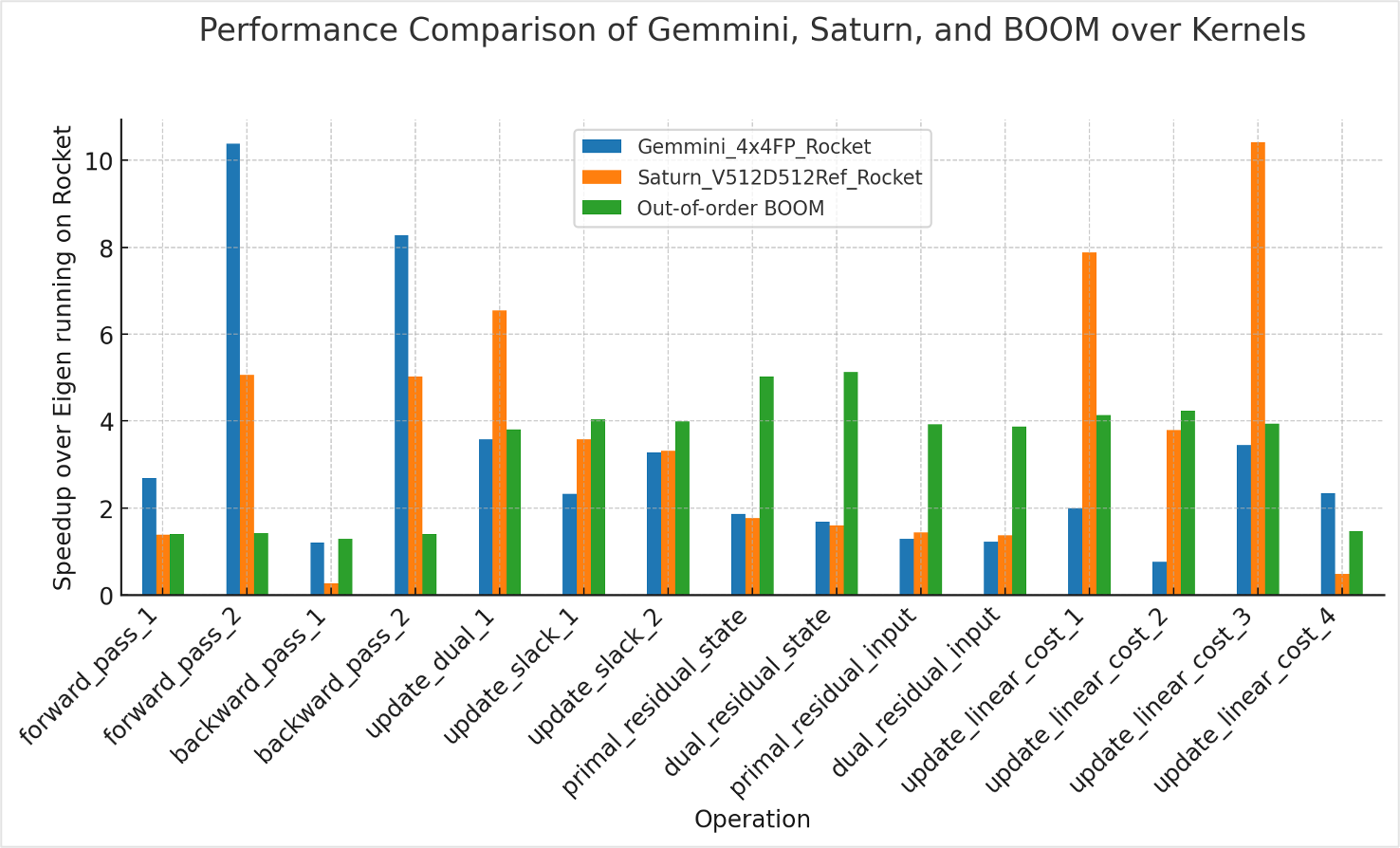}
\caption{Kernel-level performance of vector, systolic, and superscalar architectures.}
\label{fig:saturn_perf}
\vspace{-0.3cm}
\centering
\end{figure}

Due to the flexible nature of the RVV vector instruction set as well as the Saturn generator, we were able to evaluate many hardware configurations using the same software mapping. By using RVV intrinsics and dynamically computing \texttt{VLMAX}, the same binary can be executed on every hardware configuration evaluated. Since the iterative functions within TinyMPC can often only utilize four vector lanes at most, Saturn configurations with DLEN=128 are more efficient compared to their counterparts with DLEN=256. Furthermore, achieving the highest optimal performance required both a superscalar Shuttle frontend and a DLEN=256 backend. However, interestingly, the single-issue reference design outperformed the DSP and general-purpose Saturn configurations for a DLEN=256, even though both of those configurations should have outperformed the single-issue reference design. Further investigations could include evaluating more area-minimal Saturn configurations. Currently, under 1.4mm$^2$, a Rocket core is the most efficient implementation. However, minimal Saturn configurations could result in improved performance in this domain due to Saturn's instruction sequencing. For higher-performance configurations, hardware support for explicitly operating on small, dense matrices could improve Saturn's performance as the vector and datapath lengths scale beyond what was evaluated in this work.

Evaluating multiple configurations for Gemmini was more challenging, as programming Gemmini using its low-level assembly interface requires manual rewriting across changes to the mesh. Because of this, we primarily focused on creating a single, well-optimized implementation for the OSGemminiRocket configuration, which utilizes a 4$\times$4 output-stationary FP32 mesh, evaluated with both a 64KB and 32KB scratchpad. Furthermore, we include the area report for an equivalent weight stationary design, which requires the generation of a 1KB accumulator memory for computing intermediate results. However, this implementation has significantly worse performance, as software optimizations, aside from software unrolling and static mapping, have not been implemented for this design. Under an area window of 1.5mm$^2$ to 2.3mm$^2$, Gemmini is the optimal design. In this case, the efficiency improvements may be due to the ability to fuse scaling and matrix operations, as well as the fact that fine-grained instruction sequencing works well with blocks of matrices that are multiples of 4. To achieve a more thorough understanding of Gemmini's end-to-end performance characteristics, future evaluations can consider other mesh dimensions, smaller scratchpad capacities, as well as hardware modifications such as re-purposing activation and pooling functions suitable for classical control workloads.

\subsection{Hardware-in-the-Loop Evaluation}\label{sec:hil}
The end goal of optimizing robotic algorithms is to improve the performance of robot systems. In this section, we characterize the impact of our MPC optimizations in end-to-end robot scenarios.  For this evaluation, we use a hardware-in-the-loop (HIL) experimental setup.

\begin{figure}
    \centering
    \includegraphics[width=1.0\linewidth]{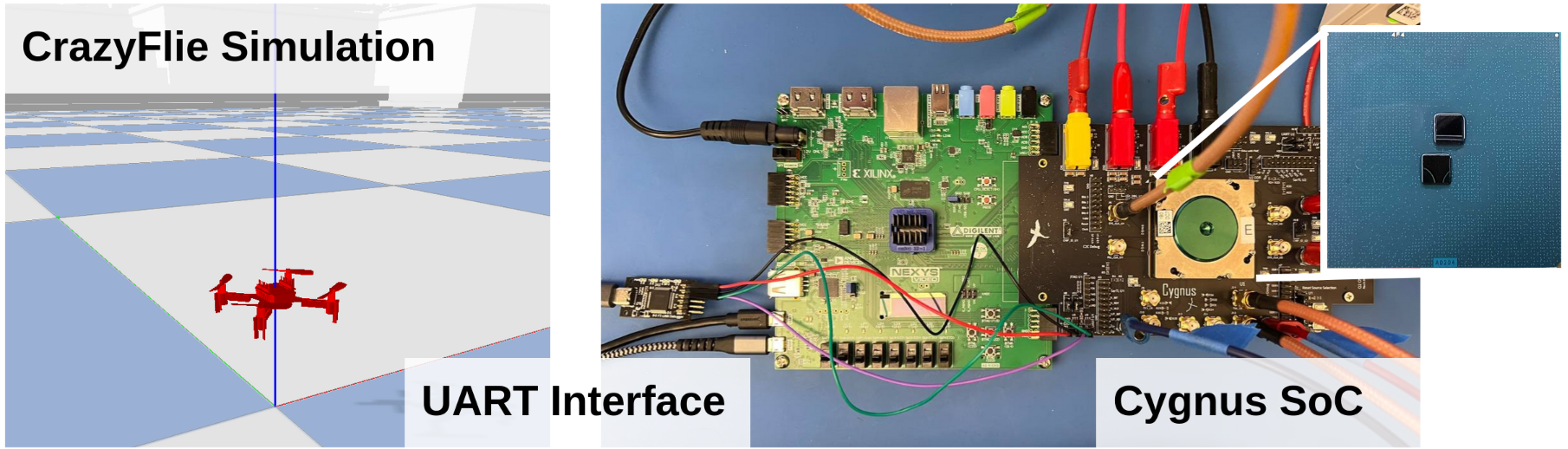}
    \caption{Experimental setup for characterizing simulated robot performance.}
    \label{fig:hil}
\end{figure}

For our experimental setup, we simulate a CrazyFlie micro-drone controlled via MPC, depicted in Figure~\ref{fig:hil}. We simulate the drone's dynamics using gym-pybullet-drones \cite{panerati2021learning}, which runs on a host computer. For our computing platform, we use Cygnus, a RISC-V vector SoC fabricated in Intel16 technology \cite{jain2025cygnus}.  We implemented a flight controller on the SoC using the Zephyr real-time operating system, using one thread to manage IO with the host PC running the simulation and another thread to execute TinyMPC. We use a UART link to transmit the drone's simulated state and desired position to the SoC. The SoC continually solves for the motor forces and transmits the solutions to the host computer over UART.

\begin{figure}[]
    \centering
    \footnotesize
    \includegraphics[width=1.0\linewidth]{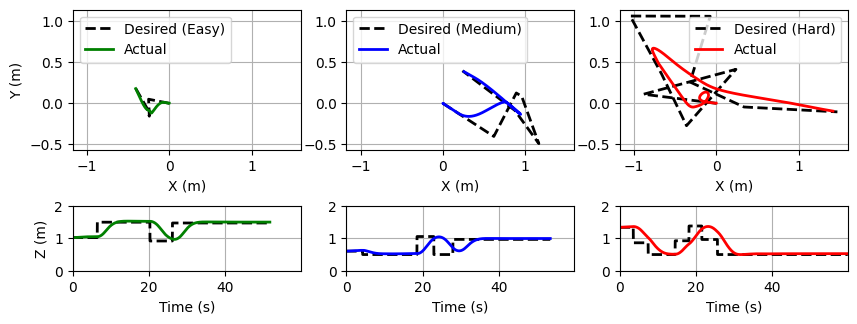}
    \begin{tabular}{|l|c|c|c|}
    \hline
         Difficulty & Easy & Medium & Hard \\ \hline
         Waypoint Count  & 5 & 7 & 10 \\
         Time Between Waypoints & 0.5s & 0.4s & 0.3s \\
         Avg. Waypoint Distance & 0.3m & 0.7m & 1.1m \\ \hline
    \end{tabular}
    \caption{Overview of scenario difficulties, along with a sample trajectory for each difficulty.}
    \label{fig:scenarios}
\end{figure}

\begin{figure}
    \centering
    \includegraphics[width=1.0\linewidth]{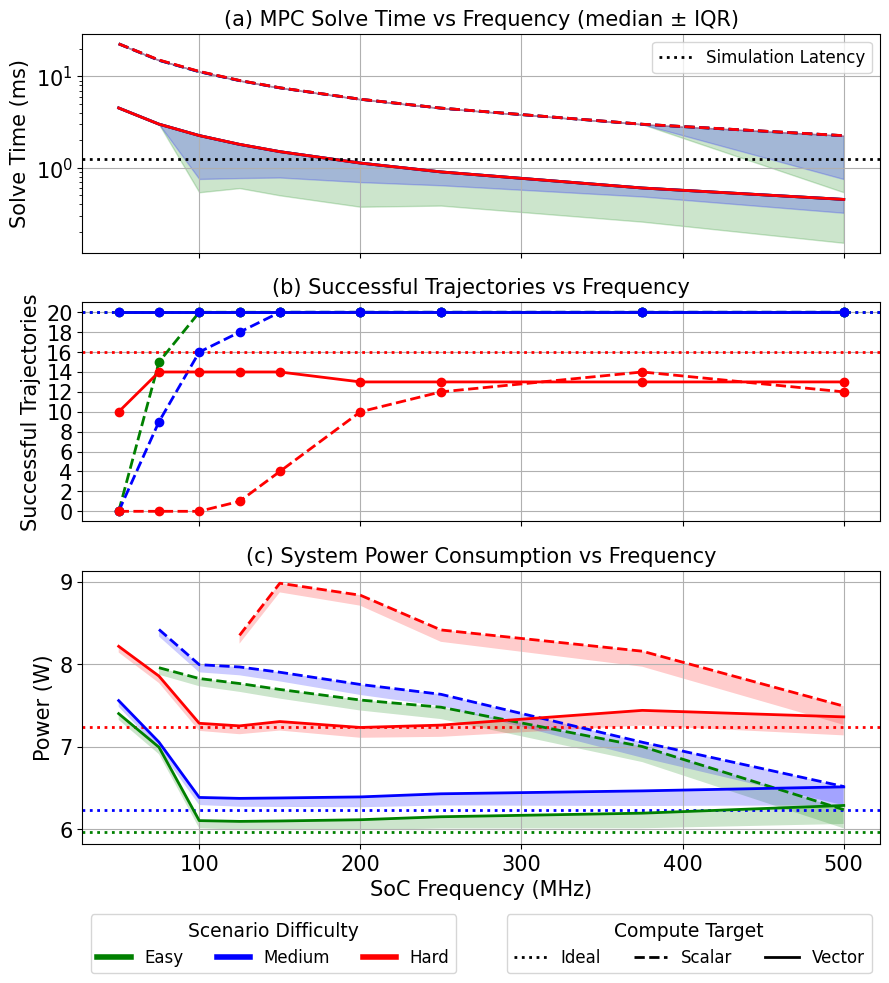}
    \caption{Impact of compute architecture and clock frequency on MPC solve time, success rate, and power consumption for drone flight scenarios.}
    \label{fig:hil-results}
\end{figure}

We evaluate the performance of the micro-drone as it tracks a series of waypoints transmitted by the host computer. The drone is not aware of future waypoints and must dynamically re-plan a trajectory upon receiving a new waypoint. We create scenarios with three difficulty categories, each featuring 20 unique sets of waypoints. As difficulty increases, so does the number of waypoints, distance between waypoints, and the frequency of waypoint changes; this information, along with an example scenario of each difficulty, is provided in Figure~\ref{fig:scenarios}.

We evaluate scalar and RVV software implementations of MPC on-chip, mapped to the large RVV core on 
our SoC, which features an in-order dual-issue superscalar shuttle core and a vector implementation with VLEN=512 and DLEN=256. We characterize the micro-drone's performance across a range of SoC clock frequencies, depicted in Figure~\ref{fig:hil-results}. The first performance metric is the solve time of MPC. We plot the median solve time along with the interquartile range; although most solves reach the maximum iteration count of TinyMPC, we observe that vector designs running at at least 100MHz and scalar designs at 500MHz can converge to a solution in fewer iterations in easy and medium difficulty scenarios. This is due to the improved ability to warm-start the solver based on prior solutions, demonstrating a compounding effect of improved computational performance that reduces solver overhead. 

The impact of the reduced solve time can be seen in Figure~\ref{fig:hil-results}-(b), where we plot how often the drone can successfully navigate to the final waypoint for each difficulty. We also plot an ideal policy, which runs MPC at each physics timestep. The ideal policy can complete every easy and medium scenario, and 80\% of the hard scenarios, which were designed to test the limits of MPC performance. The scalar baseline requires an SoC clock frequency of at least 100MHz to complete all easy tasks and 150MHz to complete all medium tasks. However, a clock frequency of at least 250MHz is needed to complete over 60\% of hard tasks. On the other hand, our vector implementation completes all easy and medium difficulty tasks at all frequencies and exceeds 60\% of hard tasks above 75MHz. We observe that real-time implementations do not match the performance of the ideal policy on hard scenarios, even when solve time falls below the latency of a simulation timestep; this is due to the additional latency introduced by UART communication. 

We also characterize the impact of the system design point on the overall power consumption of the robot platform for successfully completed tasks. The primary contribution to the power consumption of the robot is rotor power. We use a momentum-theory approximation \cite{hoffmann2007quadrotor} to model the power consumption of the simulated drone's actuators. This model is listed in Equation~\ref{eq:power}, where the induced power consumed by a rotor, $P_{ind}$, can be derived from the rotor's thrust, $T$, the propeller disk area $A$, and the air density $\rho$. Additionally, we directly measure the power consumption of the 
SoC using a bench power supply. 
\begin{equation}
    \label{eq:power}
    P_{ind} = \frac{T^{3/2}}{\sqrt{2\rho A}}
\end{equation}
We show the total power consumption of the drone in Figure~\ref{fig:hil-results}-(c), where the shaded region shows the power consumption due to compute, and the value below the shaded region represents the power consumption purely due to actuation. We also plot the power consumption of the ideal policy, which only considers actuator power. Higher difficulty tasks naturally consume more power due to the increased rate and magnitude of maneuvers the drone must perform, which is reflected in the ideal policy's performance.

At low clock frequencies, we observe substantial degradation in actuator power consumption, incurring up to 30\% overhead for scalar implementations and 22\% for vector implementations. Even if the drone can successfully complete the maneuver, we observe inefficient flight patterns such as oscillation and overcorrections at design points with under-provisioned compute resources. However, at 100MHz, our vectorized implementation can achieve within 1\% of the ideal actuator power. 

The power overhead due to compute contributes an additional 1-5\% to the system power consumption. In our experiments, scalar designs benefit from increased clock frequency across the entire frequency range due to improvements to flight patterns. However, this comes at an additional 4.5\% power overhead at 500MHz, compared to the 2\% overhead of vector designs, which achieve minimum mechanical power consumption at lower frequencies. 

\begin{figure}
    \centering
    \includegraphics[width=1\linewidth]{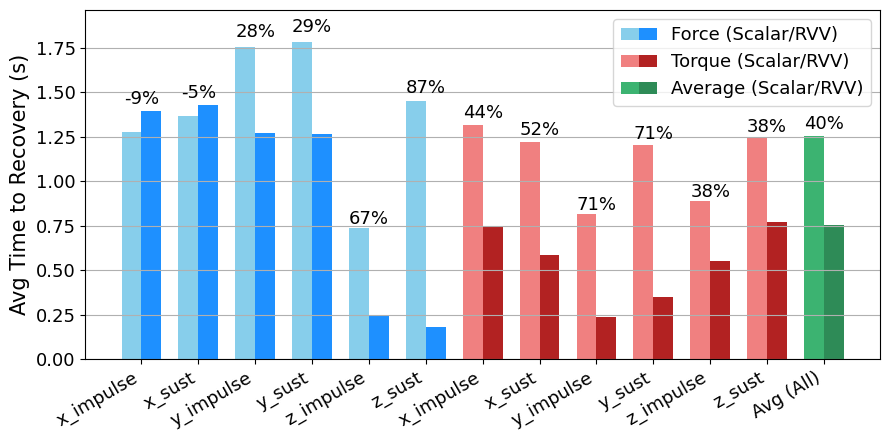}
    \caption{Impact of vectorization on disturbance recovery time. Percentage improvement is plotted above the bars.}
    \label{fig:disturbances}
\end{figure}

Finally, we evaluate the impact of hardware acceleration on robustness and distirbance rejection. We applied 100-ms step and impulse disturbances (axis-aligned forces, torques, and combined vectors) and measured the maximum recoverable magnitude and time-to-recovery (TTR; return within 5 cm for 250 ms). Our results show at 100 MHz, vectorized MPC endures 1.89× larger forces and 1.96× larger torques compared to scalar. Additionaly, the average TTR is improved by 40\%, breaking down results by disturbance category in Figure~\ref{fig:disturbances}.

\subsection{System-Level Impacts}
Robot deployments typically include multiple concurrent tasks, spanning across perception, planning, control, and reasoning. In Section~\ref{sec:hil} we characterize a system in which robot performance is not bottlenecked by control frequency. However, it is important to consider scenarios which do not directly benefit from improving control loop frequency; examples include systems with low-bandwidth sensors/actuators, or systems where end-to-end performance is limited by other tasks, such as perception.

We evaluated a scenario with two concurrent tasks: TinyMPC, and DroNet \cite{loquercio2018dronet}, a CNN used for local planning in embedded robot platforms. These tasks run on a single 100 MHz RVV core. The MPC task runs as a high-priority RTOS task at a fixed 50 Hz, while DroNet runs as a background thread. A scalar MPC implementation occupies the CPU for 28.5\% of the time, compared to only 3.3\% for a vector implementation. By swapping to the vector implementation, DroNet's framerate improves by 1.35$\times$ to 7.7 FPS, demonstrating how accelerating real-time control frees up resources for contending robotic tasks.

\subsection{Size Weight and Power Analysis}
\begin{table}[]
    \centering
    \begin{tabular}{|l|c|c|c|}
         \hline
         Drone & CrazyFlie & Hawk & Heron  \\ \hline \hline
         Specialty & Generic & Agility & Hover Efficiency \\ 
         Mass & 27 g & 46 g & 35 g \\ 
         Propeller Diameter & 45 mm & 60 mm & 90 mm \\
         Motor Arm Length & 80 mm & 80 mm & 160 mm \\
         Motor Kv &  14000 rpm/V & 28000 rpm/V & 14000 rpm/V \\
         Battery Cells & 1S & 2S & 2S \\ \hline
    \end{tabular}
    \caption{Mechanical and electrical parameters for CrazyFlie variants.}
    \label{tab:swap-params}
\end{table}

\begin{figure}
    \centering
    \includegraphics[width=\linewidth]{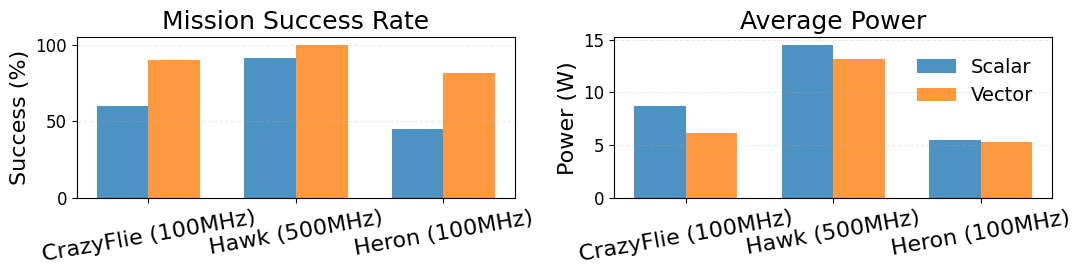}
    \caption{Mission success and power metrics for CrazyFlie variants. The clock frequency achieving lowest power consumption is used per variant.}
    \label{fig:swap-results}
\end{figure}

Robotic systems are highly sensitive to variations to their robot morphologies, especially size, weight, and power (SWaP) \cite{Neuman_2023, Krishnan_2022}. In addition to the standard CrazyFlie platform, we introduce two variants: Hawk, optimized for racing at the cost of power efficiency, and Heron, optimized for hover efficiency at the cost of agility. The specifications of these drones are listed in Table~\ref{tab:swap-params}. We generate new linearized models and policies for these drones, and evaluate them in the HIL waypoint tracking scenario.

Hawk benefits the most from hardware acceleration. Due to the combination of faster control loops and responsive actuators, Hawk can complete all waypoint tracking tasks (including hard tasks) with a 100 MHz vector implementation, a task not achieved by the baseline CrazyFlie. However, Hawk is also highly dependent  on hardware acceleration, as the scalar implementation requires at least 375 MHz to hover, and 500 MHz to complete all easy and medium tasks. Unlike the baseline CrazyFlie, Hawk continues to benefit from a higher clock speed for vectorized MPC up through 500 MHz, saving 1W of system power by clocking at 500 MHz compared to 100 MHz. On the other hand, Heron does not substantially benefit from high-frequency SoCs; similar to the baseline CrazyFlie, a 100MHz vectorized implementation is sufficient for achieving the best actuator power consumption.

A summary of success and power metrics for the best-performing SoC frequencies for each design is shown in Figure~\ref{fig:swap-results}. These SWaP results suggest high-performance SoCs for Hawk, where compute cost is small relative to actuation, and low-power, custom hardware for Heron’s sluggish dynamics which do not benefit as much from fast control loops and which consume less actuator power.

\section{Conclusion}
Our profiling and optimization across various computing platforms, ranging from CPUs and vector machines to domain-specialized accelerators, encompass both kernel-level benchmarks and a comprehensive end-to-end robotic workload. We compare integrated accelerators, such as RISC-V cores with vector extensions, to more decoupled systems, like systolic arrays, evaluating their performance, area efficiency, and utilization. This exploration not only quantifies the trade-offs inherent between these architectural paradigms but also underscores how the choice of hardware architecture is contingent upon specific workload characteristics and application demands. In particular, we observe that hardware schedulers, such as instruction sequencers in vector machines and CISC instructions for co-processors, have minimal benefit in real-time control workloads. However, we find that we can achieve significant performance gains through careful software scheduling, resulting in up to 3.7$\times$ speedup over baseline implementations. 

We identify that a significant challenge for deploying optimized robotics algorithms on specialized hardware is the engineering overhead of hand-optimizing the software mapping. To facilitate rapid evaluation of new hardware designs, we pursue two potential options. First, we explore a library-based approach supporting common operations. Existing interfaces, such as Eigen \cite{eigenweb}, result in poor utilization for small operands and do not provide clean abstractions for specialized memory units such as software-managed scratchpads. We contribute the \texttt{matlib} library, which provides a simple programming interface for embedded optimization, as well as automated optimization passes that apply relevant optimizations we identified in our characterization. Interesting future directions include representations that directly express robotic workloads as computation graphs, enabling the use of established compilation infrastructures, such as MLIR \cite{lattner2021mlir}.

Finally, we explore how accelerating robot control can improve the performance of other concurrent tasks in a robot system. Given optimized mappings of multiple algorithms to several hardware backends, future work will consider strategies for scheduling robotic applications to heterogeneous embedded SoCs while addressing physical and computational constraints in robotic systems.

\section*{Acknowledgment}

This research was supported in part by the NSF Awards CCF-
2238346 and in part by SLICE Lab industrial sponsors and affiliates. This project is funded by the Microelectronics Commons Program, a DoD initiative. Distribution Statement A: Approved for public release. Distribution is unlimited. The views and conclusions contained in this document are those of the authors and should not be interpreted as representing the official policies, either expressed or implied, of the U.S. Government.

\newpage
\twocolumn[
\vspace{1em}
]

\newpage
\newpage
\bibliographystyle{IEEEtranS}
\bibliography{main/sophia, main/refs}
\newpage 
\null 
\newpage

\appendix
\section{Artifact Appendix}

\subsection{Abstract}
This artifact appendix describes how to use Chipyard to run an RTL experiment for Accelerated-TinyMPC; as well as run hardware-in-loop experiments with end-to-end quadcopter simulations, and how to reproduce results as shown in Section~\ref{evaluation}. 

\subsection{Artifact Meta-Information Checklist}
{\small
\begin{itemize}
    \item {\bf Runtime environment:} Ubuntu 24.04 LTS 
    \item {\bf Hardware:} An x86-64 CPU. This work used an AMD Ryzen Threadripper PRO 5965WX. Hardware-in-the-loop (HIL) experiments use a Nexys Video FPGA and the Cygnus research chip\cite{jain2025cygnus}.

    \item {\bf Required Disk Space:} 50GB
    \item {\bf Experiments:} Experiment 1: RTL Evaluation of TinyMPC under varying SoC configurations. Experiment 2: HIL experimentation with a test chip and a drone simulator.
    \item {\bf Languages:} C/C++ (TinyMPC/matlib/RTOS), Python (Scripting, Codegen), Chisel (RTL).
    \item {\bf Quantitative Metrics:}  Part 1: End-to-end solve latency and kernel latency (cycles). Part 2: Solve time, mission success rate, power consumption.
    \item {\bf Qualitative Metrics:} Flight trajectories, flight recordings.
    \item {\bf Output:} RTL simulation outputs, UART transmissions from test chip, recordings and renderings from the simulated drone.
    \item {\bf Installation Time:} 1 hour (scripted installation).
    \item {\bf Experiment Duration:} 4 hours.
    \item {\bf Publicly available:} Yes.
    \item {\bf Code licenses:} Several, see download.

\end{itemize}
}

\subsection{Description}
\subsubsection{How to access}
This artifact includes the following open-source repositories. Instructions for installation are provided in the experimental procedures. Artifacts may be obtained from github or from Zenodo at \url{https://doi.org/10.5281/zenodo.17065844}.
\begin{enumerate}
    \item Accelerated-TinyMPC: An implementation of TinyMPC for scalar, vector, and systolic architectures (\url{ https://github.com/ucb-bar/Accelerated-TinyMPC/tree/iiswc-ae}) 
    \item matlib: A hardware-accelerated RISCV matrix library (\url{https://github.com/wid4soe/matlib/tree/iiswc-ae}), used in Accelerated-TinyMPC.
    \item zephyr-chipyard-software: Zephyr-RTOS-based build environment for deployment to RISCV prototype SoCs (\url{https://github.com/ucb-bar/zephyr-chipyard-sw/tree/iiswc-ae}).

\end{enumerate}

\subsubsection{Dependencies - Hardware} 
Experiment 1 does not require any specialized hardware. However, Experiment 2 requires access to a custom research chip in order to run the HIL experiments. To facilitate evaluation of this experiment, the authors will provide SSH access to the host PC connected to the chip.

\subsubsection{Dependencies - Software} 
This artifact uses miniforge (\url{https://github.com/conda-forge/miniforge}) to manage software dependencies. All software dependencies are open-source/public, and have been tested on Ubuntu 24.04.

\subsection{Installation}
Full installation instructions are provided in the experiment workflow sections, as installation instructions vary depending on which experiment is being performed.

\subsection{Experiment Workflow}
Running all the steps below in a \texttt{screen} or \texttt{tmux} session is recommended to account for long-running experiments.
\subsubsection{Experiment I: Running Accelerated-TinyMPC with Simulated Hardware}

This experiment will build Accelerated-TinyMPC for scalar, vector, and systolic backends, and evaluate the performance on RTL implementations.

\paragraph{Cloning and Installation:} Clones the main repository, installs dependencies via miniforge, and clones and installs Chipyard.
\begin{lstlisting}[style=bashstyle]
$ cd ~
$ git clone -b iiswc-ae \
https://github.com/ucb-bar/Accelerated-TinyMPC.git
$ cd Accelerated-TinyMPC
$ git submodule update --init
$ source scripts/install_conda.sh
$ bash scripts/install_chipyard.sh
$ source tools/chipyard/env.sh
\end{lstlisting}

\paragraph{Building Binaries:} The following script will use CMake to build baremetal binaries. The compiled binaries for each target are exported to \texttt{build-[target]}, which measure end-to-end runtime, and \texttt{build-[target]-cycles}, which measure kernel-level runtimes.
\begin{lstlisting}[style=bashstyle]
$ bash scripts/build_all.sh
\end{lstlisting}
\paragraph{RTL Simulation:} The following script runs RTL simulations using the previously generated binaries. Chipyard configurations are compiled to Verilog sources using Chisel, and RTL simulation is performed using Verilator. The script prints UART output from the simulated SoC to \texttt{stdout}. Note: This step may take a couple of hours to complete.
\begin{lstlisting}[style=bashstyle]
$ bash scripts/run_rtl.sh
\end{lstlisting}

\paragraph{Plotting:} Plots end-to-end solve performance and kernel runtimes to \texttt{results/plots}. Raw UART outputs are stored to \texttt{results/rtl}.
\begin{lstlisting}[style=bashstyle]
$ python3 scripts/plot_results.py
\end{lstlisting}

\subsubsection {Experiment II: HIL Experiment for Accelerated-TinyMPC}
\begin{figure}
    \centering
    \includegraphics[width=0.9\linewidth]{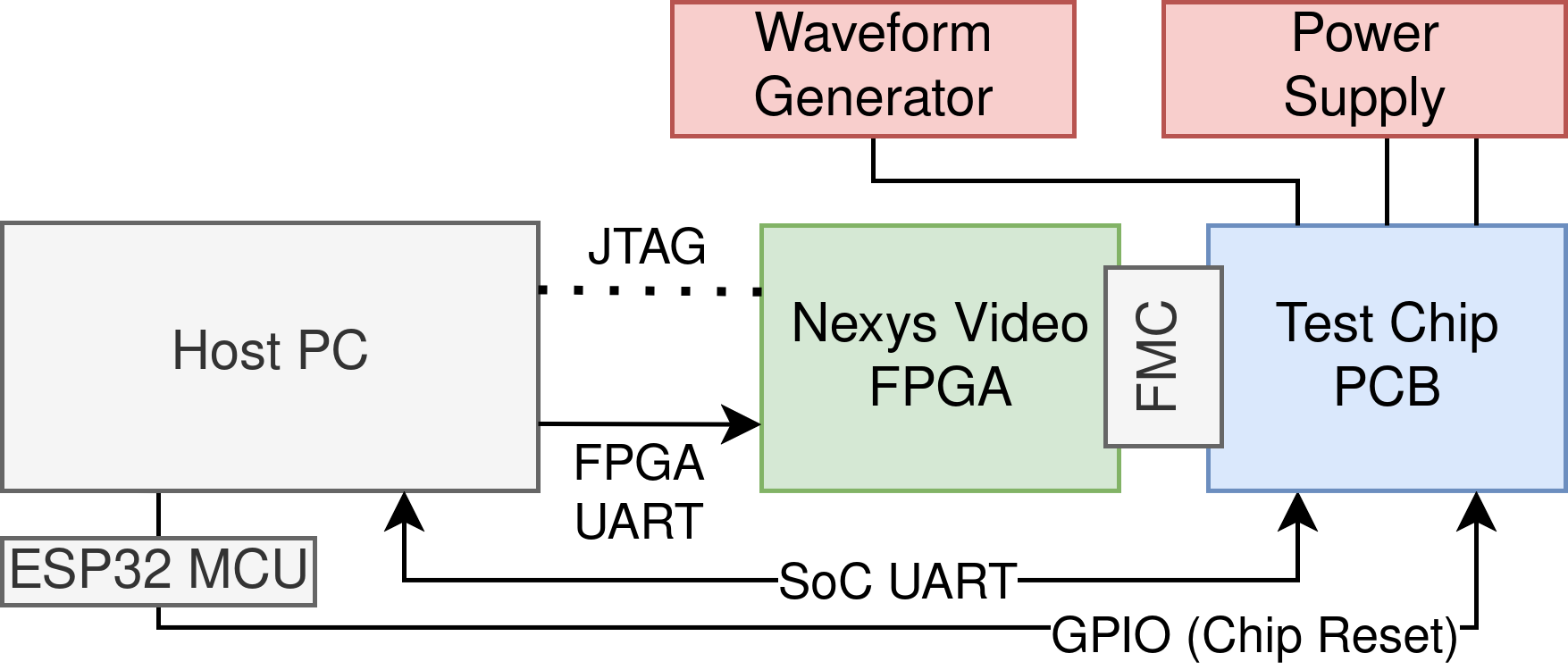}
    \caption{HIL Experiment Block Diagram}
    \label{fig:hil-ae}
\end{figure}

The following section documents how to run the experiment on the Cygnus SoC used in the paper; To target a different SoC, users should use a different Zephyr board config. The experimental setup is depicted in Figure~\ref{fig:hil-ae}. The SoC is tethered to the host platform with a Nexys Video FPGA, used for uploading program binaries via UART and for backing memory. Additionally, the SoC's UART is directly connected to the host PC using a USB-UART adapter. An ESP32 Microcontroller is used to reset the chip via GPIO. The host PC also runs a quadcopter simulator. 

\paragraph{Installation:} Clone the \texttt{zephyr-chipyard-sw} repository, which targets the SoC as a Zephyr RTOS board. Installs Zephyr and gym-pybullet-drones. Next, this compiles GCC 15.1.0 (needed to support RVV intrinsics), which has been patched for compatibility with the Zephyr framework.
\begin{lstlisting}[style=bashstyle]
$ cd ~
$ git clone -b iiswc-ae \
https://github.com/ucb-bar/zephyr-chipyard-sw.git
$ cd zephyr-chipyard-sw
$ source scripts/install_conda.sh 
$ source scripts/install_submodules.sh
$ bash scripts/install_toolchain.sh
$ source scripts/set_envvars.sh
\end{lstlisting}

\paragraph{Build Scalar Controller:} Use the following command to build a scalar implementation to \texttt{build/zephyr/zephyr.elf}.
\begin{lstlisting}[style=bashstyle]
$ west build -b chipyard_cygnus samples/drone_control/ -p
\end{lstlisting}

\paragraph{Reset and Flash SoC:} First, send a command to reset the SoC to enable uploading program binaries. Next, use \texttt{pyuartsi} to upload the binary via UART. \textbf{Important Note:} Always reset the SoC before uploading a binary; otherwise, the setup might need manual power cycling by the authors.
\begin{lstlisting}[style=bashstyle]
$ python3 scripts/reset_soc.py
$ bash scripts/flash_chip.sh
\end{lstlisting}

\paragraph{Run HIL Simulation:} Run a headless pybullet simulation. This simulation models an SoC running at 100MHz. To accommodate the high UART latency via Python, the simulation is executed at 0.2x speed, and the SoC is clocked at 20MHz. Note: this step may take 30-60 minutes. 
\begin{lstlisting}[style=bashstyle]
$ bash scripts/run_pybullet.sh data/scalar
\end{lstlisting}

\paragraph{Repeat with Vector Code:} Re-run the prior script with a vectorized implementation.
\begin{lstlisting}[style=bashstyle]
$ west build -b chipyard_cygnus samples/drone_control/ \ 
-p -- -DRISCV_VECTOR=ON
$ python3 scripts/reset_soc.py
$ bash scripts/flash_chip.sh
$ bash scripts/run_pybullet.sh data/vector
\end{lstlisting}

\paragraph{Render Videos and Plot Results:} First, plot the results, exported to \texttt{results/plots}. Next (optional), render videos to visualize the drones' behaviors, saved to \texttt{results/videos}. 
\begin{lstlisting}[style=bashstyle]
$ python3 scripts/plot_results.py
$ bash scripts/render_videos.sh
\end{lstlisting}

\paragraph{Regenerate Figure~\ref{fig:hil-results}:} The work's original HIL data are provided in full, and the plots can be regenerated as follows if accessing via Zenodo:
\begin{lstlisting}[style=bashstyle]
$ python3 scripts/plot_hil_data.py
\end{lstlisting}

\subsection{Interpreting Results}
\subsubsection{Experiment I}
This experiment runs a subset of the measurements in this work, due to the long runtime of RTL simulation. The selected design points intend to cover the scalar, vector, and systolic architectures studied. Scalar code includes both matlib and Eigen code. Vector code includes the original matlib implementation, as well as the final hand-optimized implementations. Systolic code includes the final output-stationary optimized Gemmini code. The results are executed on two scalar cores, two vector cores, and one systolic core.

The results should be comparable to the end-to-end and kernel-level performances reported in the paper. However, some variations may occur; RVV results are expected to improve, as these experiments run using the latest implementation of the Saturn RVV core, while the reported results are on the older version of Saturn used during SoC tapeout, which is currently not part of an official release.

\subsubsection{Experiment II} The results of the HIL experiment should be close to the results in Figure~\ref{fig:hil-results} for vector and scalar code at 100MHz. Slight variation may occur for different host platforms, which may impact Pybullet simulation performance and IO latency.
\subsection{Experiment Customization}
Users may modify hardware configurations in Experiment~1's \texttt{run\_rtl.sh} to cover more design points. Furthermore, although the \texttt{example\_quadrotor\_tracking(\_rvv).cpp} samples have been shortened to 1 iteration to reduce RTL simulation time, this can be increased back to 10. For Experiment 2, the reviewers may try other configurations in \texttt{samples/drone\_control/CMakeLists.txt}, such as enabling unoptimized RVV code. If familiar with Zephyr, they can modify other parameters, such as UART baudrate or target board.

\subsection{Methodology}

Submission, reviewing and badging methodology:

\begin{itemize}
  \item \url{https://www.acm.org/publications/policies/artifact-review-and-badging-current}
  \item \url{https://cTuning.org/ae}
\end{itemize}

\end{document}